\documentclass[pdflatex,sn-nature,twocolumn]{sn-jnl}
\makeatletter
\@twosidefalse
\@mparswitchfalse
\makeatother

\usepackage{graphicx}%
\usepackage{multirow}%
\usepackage{amsmath,amssymb,amsfonts}%
\usepackage{amsthm}%
\usepackage{mathrsfs}%
\usepackage[title]{appendix}%
\usepackage{xcolor}%
\usepackage{textcomp}%
\usepackage{manyfoot}%
\usepackage{booktabs}%
\usepackage{algorithm}%
\usepackage{algorithmicx}%
\usepackage{algpseudocode}%
\usepackage{listings}%
\usepackage{algcompatible}%
\usepackage{flushend}    
\usepackage{tabularx}
\usepackage{comment}

\theoremstyle{thmstyleone}%
\theoremstyle{thmstyletwo}%

\theoremstyle{thmstylethree}%

\begin{document}

\twocolumn[
  \vspace{1cm}
  \begin{center}
  {\Large\bfseries Coding-Free and Privacy-Preserving Agentic Framework for Data-Driven Clinical Research}
  \end{center}
  
  \begin{center}{\large
  Taehun Kim\textsuperscript{*1,2},
  Hyeryun Park\textsuperscript{*3},
  Hyeonhoon Lee\textsuperscript{*3,4,5},
  Yushin Lee\textsuperscript{1},\\
  Kyungsang Kim\textsuperscript{\dag 6},
  Hyung-Chul Lee\textsuperscript{\dag 1,3,4}
  }
  \end{center}
  
  {\normalsize\raggedright
  \textsuperscript{1}Infmedix, Co., Ltd., Seoul, Republic of Korea.\\
  \textsuperscript{2}Department of Transdisciplinary Studies, Seoul National University, Seoul, Republic of Korea.\\  
  \textsuperscript{3}Healthcare AI Research Institute, Seoul National University Hospital, Seoul, Republic of Korea.\\
  \textsuperscript{4}Department of Medicine, Seoul National University College of Medicine, Seoul, Republic of Korea.\\
  \textsuperscript{5}Department of Transdisciplinary Medicine, Seoul National University Hospital, Seoul, Republic of Korea.\\
  \textsuperscript{6}Department of Radiology, Massachusetts General Hospital and Harvard Medical School, Boston, MA, USA.\\
  \textsuperscript{*}These authors contributed equally to this article.\\
  \textsuperscript{\dag}Co-correspondence: Kyungsang Kim (kkim24@mgb.org), Hyung-Chul Lee (vital@snu.ac.kr).
  \par}
  
  \vspace{0.5cm}
  \begin{center}
  {\large\bfseries Abstract}
  \end{center}
  Clinical data-driven research requires clinical expertise, programming skills, access to patient data, and extensive documentation, creating barriers and slowing the pace for clinicians and external researchers.
  To address this, we developed the Clinical Agentic Research Intelligence System (CARIS) that automates the workflow: research planning, literature search, cohort construction, Institutional Review Board (IRB) documentation, Vibe Machine Learning (ML), and report generation, with human-in-the-loop refinement.
  CARIS integrates Large Language Models (LLMs) with modular tools through the Model Context Protocol (MCP), enabling natural language–driven research without coding while allowing users to access only outputs.
  We evaluated CARIS on three heterogeneous datasets with distinct clinical tasks, where it completed planning and IRB documentation within four iterations, supported Vibe ML, and generated reports, achieving 96\% completeness in LLM-based evaluation and 82\% in human evaluation.
  CARIS demonstrates potential to reduce documentation burden and technical barriers, accelerating data-driven clinical research across public and private data environments.
  \vspace{1cm}
]

\section*{Introduction} 

\begin{figure*}[ht]
    \centering
    \includegraphics[width=0.8\textwidth]{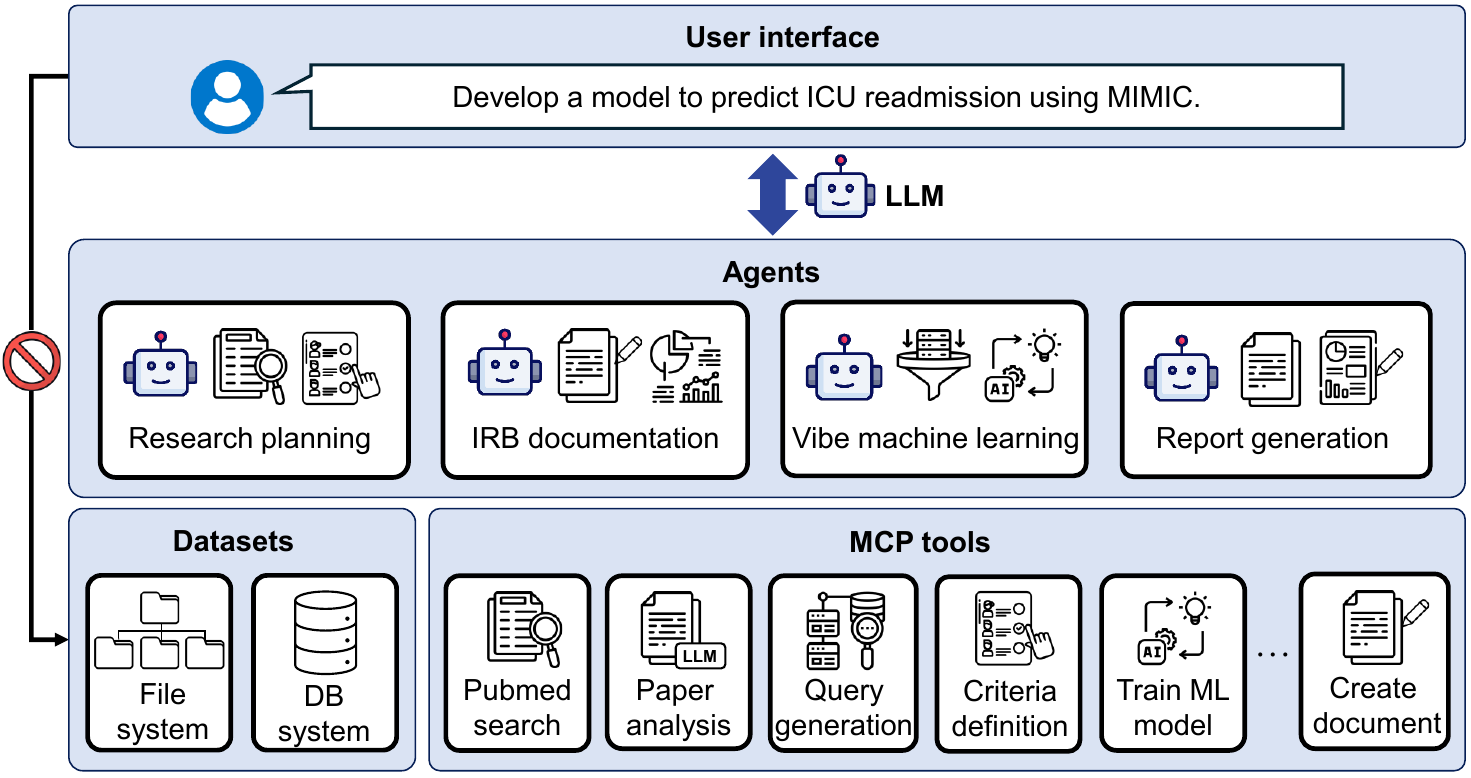}
    \caption{\textbf{Overview of Clinical Agentic Research Intelligence System (CARIS).} Given a user intent, CARIS autonomously executes a multi-step clinical research workflow by interacting with agents, heterogeneous datasets, and various tools, while allowing user intervention. Users do not have direct access to the file or DB systems.
    }
    \label{overview}
\end{figure*} 

Clinical data-driven research requires domain expertise, programming skills, and access to sensitive patient data to translate clinical questions into actionable insights. 
Although clinicians formulate hypotheses based on their clinical knowledge, they often encounter substantial barriers in data preprocessing, Machine Learning (ML) model development, and result interpretation, which are time-consuming and technically demanding \cite{watson2020overcoming, thirunavukarasu2023democratizing}. 
External researchers face additional barriers in accessing private and sensitive patient data. 
These constraints impede broader participation in clinical research and slow the pace of medical advancement.

Recent advances in Large Language Models (LLMs) \cite{Touvron2023, Chowdhery2023, Meta:2025:LLaMA4, Gemma3:2025} and agentic AI systems  \cite{wang2024survey, li2024survey, zhao2025llm, wu2024autogen, tran2025multi} have demonstrated the potential to automate complex, multi-step workflows.
LLMs excel at interpreting natural language instructions and performing reasoning, while agentic systems enable them to orchestrate modular tools to execute specific tasks. 
These approaches have shown the ability to translate user intent into structured computational subtasks in general scientific domains \cite{baek2025researchagent, guo2024ds, gridach2025agentic, swanson2024virtual}, highlighting their potential for clinical research. 
Although recent efforts have made progress in isolated components, including automated ML, protocol drafting, and document generation, they still struggle to support integrated clinical workflows and secure data execution \cite{artsi2025large}. 

The existing approaches remain limited by challenges in handling heterogeneous data, maintaining data privacy, and ensuring transparency, reproducibility, and interpretability \cite{hassan2024barriers, garcia2025multi, lee2025advancements, holmes2021electronic, ehtesham2025enhancing}.
Essential components, such as Institutional Review Board (IRB) documentation and structured research protocol development, which are prerequisites for data access and ethical approval, remain insufficiently supported. 
As a result, these systems struggle to support integrated clinical workflows and face challenges to reflect the regulatory requirements and data environments specific to clinical research. 

The Model Context Protocol (MCP) addresses these limitations by providing a structured and verifiable interface between LLMs and external tools \cite{singh2025survey, luo2025mcp, attrach2025conversational, ehtesham2025enhancing}. 
It introduces a client–server architecture that decouples user interaction from underlying data sources, ensuring that only processed results are returned for review.
Within this framework, LLMs can invoke modular tools, including APIs, functions, and datasets, without direct access to sensitive patient data.
This design enables research across heterogeneous datasets while improving compliance with data governance.

This study presents the Clinical Agentic Research Intelligence System (CARIS), an agentic AI framework that automates end-to-end clinical research workflows from natural language input, as shown in Figure~\ref{overview}.
CARIS interprets user-defined research intent and orchestrates a series of agents, including research planning, Institutional Review Board (IRB) documentation, Vibe ML, and report generation, without requiring coding or direct data access.
By enabling clinicians to conduct studies through a Vibe ML paradigm, CARIS lowers technical barriers, while its MCP-based architecture ensures privacy-preserving interaction with heterogeneous data sources.
The system also alleviates the burden of clinical and regulatory documentation, including IRB preparation and report generation.
Across three distinct datasets and clinical tasks, CARIS demonstrated generalizability by completing research planning and IRB documentation within three to four iterations, supporting Vibe ML with automated visualization, and generating reports achieving high coverage (96\% by LLM vs. 82\% by human evaluation) based on a checklist derived from TRIPOD+AI \cite{collins2024tripod}.

\section*{Methods}  

\subsection*{Research Workflow Automation}

CARIS autonomously executes end-to-end clinical research workflows through natural language interaction, leveraging LLMs to orchestrate multi-step processes through structured tool invocation (Note 1 in the Supplementary).
This orchestration prompt guides the LLM to act as an independent clinical research scientist, leading study design, data analysis, tool utilization, documentation, and critical thinking.
It utilizes modular MCP tools, each performing a single atomic operation with well-defined inputs and structured outputs.
This design enables auditing of individual tools and integration of additional tools for scalability and flexibility.
The complete set of tools is provided in Table S1 (Note 2 in the Supplementary).

\begin{figure*}[t]
    \centering
    \includegraphics[width=\textwidth]{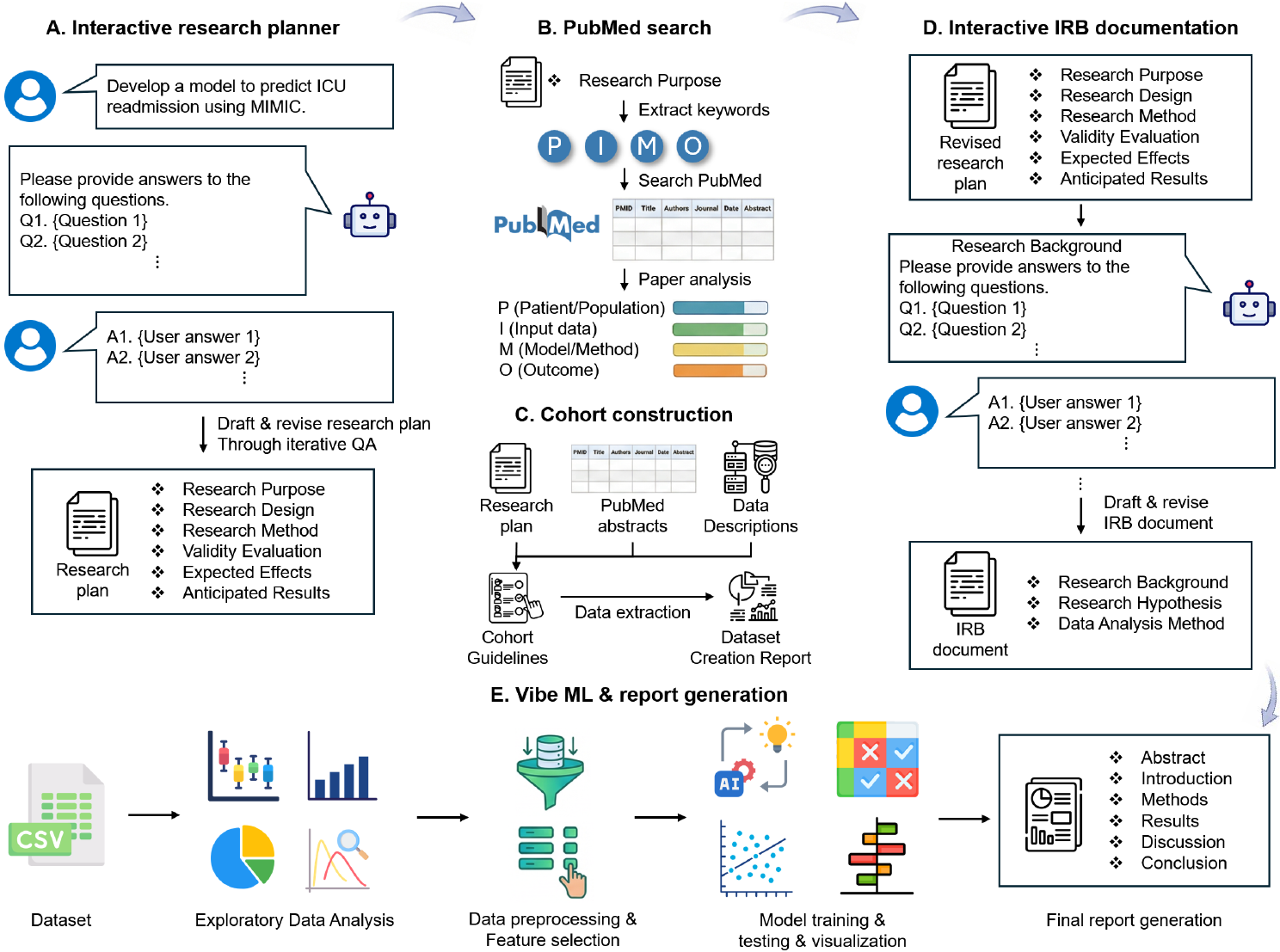}
    \caption{\textbf{Overview of the clinical research workflow.} (A) User interaction to iteratively generate and refine a structured research plan. (B) Keyword extraction using the PIMO framework and retrieval of relevant PubMed literature. (C) Database exploration, cohort definition, query execution, and dataset generation. (D) Iterative generation and refinement of IRB document sections through user interaction. (E) Data analysis, preprocessing, model training, evaluation, and visualization, followed by the automated generation of a clinical ML report.}
    \label{entire_workflow}
\end{figure*} 

\subsection*{Modular Agents}

\subsubsection*{\textit{Research Planning Agent}}
Given a user research topic, the agent automatically generates a research plan through an interactive process, as shown in Figure~\ref{entire_workflow}A (Note 3 in the Supplementary).
The agent initiates a chatbot that presents questions to progressively refine the plan.
As the user responds, the agent drafts the title, research purpose, research design, research method, validity evaluation, expected effects, and anticipated results, which are sections typically included in IRB documents.
Users can also request revisions to the generated content.

The agent supports users by retrieving and analyzing the relevant PubMed articles, as illustrated in Figure~\ref{entire_workflow}B (Note 4 in the Supplementary).
For this purpose, keywords are extracted from the research plan using the PIMO framework, a variant of the PICO framework \cite{brian1997pico} tailored for clinical ML research.
The PIMO framework captures key study elements: P (Patient/Population), I (Input data), M (Model/Method), and O (Outcome).
Based on combinations of these keywords, relevant articles are retrieved using best match sorting \cite{fiorini2018best}, along with structured metadata such as PubMed identifier (PMID), title, and abstract.
The agent compares each article against the proposed study across PIMO dimensions using 0-50 scoring and a three-stage pipeline consisting of rule-based filtering, binary screening, and fine-grained similarity scoring, producing similarity scores and brief rationales.
Based on the top ten relevant articles, users can critically review and refine their research plan.

\begin{table*}[ht]
\centering
\caption{Summary of datasets and target tasks. RDB: Relational Database.}
\resizebox{\textwidth}{!}{
\begin{tabular}{@{}lllll@{}}
\hline
Dataset & Data source & Data format & Prediction task & Number of instances \\
\hline
MIMIC-IV & Beth Israel Deaconess Medical Center & RDB & ICU readmission & 83,101 \\
INSPIRE & Seoul National University Hospital & RDB & Postoperative AKI & 100,474 \\
SyntheticMass & Synthetic Massachusetts patient data & OMOP-CDM & Prediabetes-to-diabetes & 2,556 \\
\hline
\end{tabular}
}
\label{data_task_statistics}
\end{table*}

Finally, the agent reviews the research plan and the retrieved PubMed metadata to proceed with cohort construction, as shown in Figure~\ref{entire_workflow}C (Note 5 in the Supplementary).
CARIS enables users to explore heterogeneous predefined clinical datasets.
During metadata exploration, the agent analyzes the schema to understand the data structure, including tables, fields, relationships, and semantic descriptions.
Based on this information, it generates a cohort construction guideline defining the study population (inclusion/exclusion criteria), target variable definitions, information on other variables, and key queries.
The agent then executes SQL queries to extract the required data, stores the results in a unified CSV file, and produces a dataset creation report.

\subsubsection*{\textit{IRB Documentation Agent}}

Based on the refined research plan, the same interactive chatbot is used to further generate research background, data analysis methods, and research hypotheses sections, as shown in Figure~\ref{entire_workflow}D.
Throughout this process, the agent leverages LLMs and predefined instructions to ensure that each section is generated coherently while adhering to ethical requirements, forming a complete IRB document (Note 6 in the Supplementary).
An iterative human-in-the-loop protocol allows users to continuously review and revise generated content through the chat interface.

\subsubsection*{\textit{Vibe Machine Learning Agent}}

Given the input CSV data and generated documents, the agent performs Exploratory Data Analysis (EDA) and Vibe ML, as shown in Figure~\ref{entire_workflow}E (Note 7 in the Supplementary).
During EDA, it computes descriptive statistics, assesses data quality, and generates visualizations such as distribution plots, correlation heatmaps, and pair plots.
Data preprocessing includes label-encoding categorical variables, removing columns with more than 50\% missing valuess, and mean/mode imputation for numerical and categorical variable.
Feature selection is performed based on user preference, using methods such as Recursive Feature Elimination (RFE), Boruta with SHapley Additive exPlanations (SHAP), SelectKBest using Mutual Information (MI), Random Forest–based feature importance, or LLM-recommended features.

Following this, the agent trains ML models, while allowing users to adjust features and model choices through interactive chat.
Supported algorithms are Random Forest, XGBoost, LightGBM, CatBoost, Decision Tree, Extra Trees, and linear models, with hyperparameter tuning via grid search and stratified 5-fold cross-validation.
Model performance is evaluated and visualized using confusion matrices, precision–recall curves, and AUROC with 95\% confidence intervals, with statistical comparisons based on the DeLong test and bootstrap resampling (n = 2,000, $p < 0.05$).
Model interpretability is supported by SHAP-based feature importance analyses \cite{lundberg2017unified}.

\subsubsection*{\textit{Report Generation Agent}}

The report generation agent automatically drafts a manuscript using all materials within the project.
To support structured writing of clinical ML report, it employs a prompt based on the TRIPOD+AI guideline \cite{collins2024tripod}, a standardized reporting checklist (Note 8 in the Supplementary).
The final manuscript is produced in Word format and includes all standard sections: Title, Abstract, Introduction, Methods, Results, Discussion, Conclusion, References, and Supplementary Materials, along with tables and figures.

\begin{figure*}[h]
    \centering
    \includegraphics[width=\textwidth]{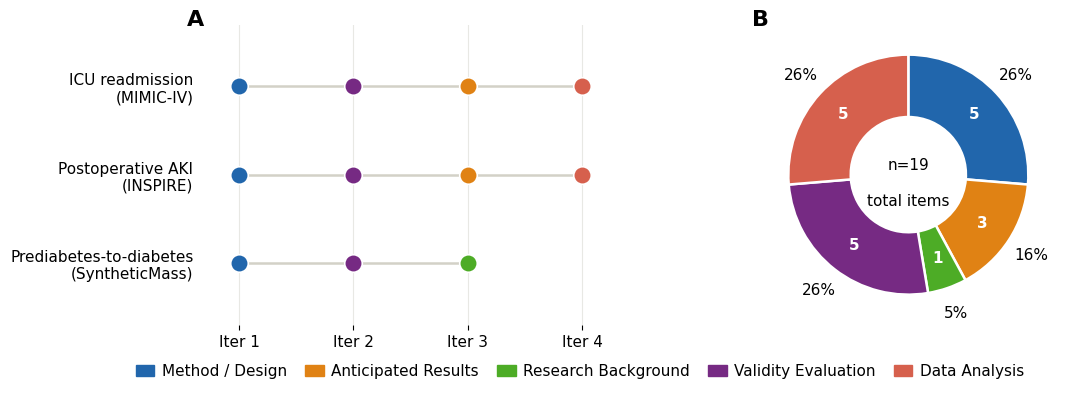}
    \caption{\textbf{Revision patterns in IRB document generation.} (A) Sections revised across iterations. Each dot represents a revision, with colors indicating section types. (B) Distribution of revision items per section. Numbers inside wedges indicate item counts, and percentages indicate their proportions.}
    \label{IRB_documentation_revision}
\end{figure*} 

\subsection*{Datasets and Experimental Settings}

To evaluate the proposed CARIS, we conduct experiments using three publicly available datasets with distinct clinical tasks: Intensive Care Unit (ICU) readmission prediction using the Medical Information Mart for Intensive Care (MIMIC)-IV dataset \cite{johnson2023mimic, mimic-iv-physionet}, postoperative acute kidney injury (AKI) prediction using the INformative Surgical Patient dataset for Innovative Research Environment (INSPIRE) dataset \cite{lim2024inspire, inspire-physionet}, and prediabetes-to-diabetes progression prediction using the SyntheticMass \cite{synthea-physionet}.
Table~\ref{data_task_statistics} summarizes the datasets and tasks, and details, including dataset versions, outcome definitions, inclusion/exclusion criteria, and predictor variables are provided (Note 9 in the Supplementary).

CARIS is implemented as a web-based application, with an LLM-driven frontend (NiceGUI) for user interaction and a backend for tool execution (Note 10 in the Supplementary).
The system supports multiple LLM APIs for flexible deployment under varying constraints.
All experiments in this study use Claude Sonnet 4.6 via API without GPUs.
The implementation, results, and demo will be available on GitHub: \url{https://github.com/thkim107/nocode_clinical_ai}.

\section*{Results} 

Across three heterogeneous datasets, CARIS autonomously executed a multi-step research pipeline.
To assess system reliability and output quality, we evaluated key results produced throughout the workflow.

\subsection*{IRB Document Evaluation}

The IRB document generation process followed a human-in-the-loop framework, where experts iteratively reviewed and revised the outputs.
The interaction between experts and the system was analyzed to assess the number and types of revisions required for completion (Note 11 in the Supplementary).
As shown in Figure~\ref{IRB_documentation_revision}A, documents were completed within three to four iterations: ICU readmission and postoperative AKI required four iterations, while the prediabetes-to-diabetes task required three.
As multiple items are often revised per iteration, a total of 19 item-level revisions were identified across all tasks, as shown in Figure~\ref{IRB_documentation_revision}B, primarily in study design and methodology (26\%), data analysis methods (26\%), and validity evaluation (26\%) sections.

Revisions in study design and methods focused on refining outcome definitions, inclusion/exclusion criteria, and alignment with dataset characteristics.
Validity evaluation revisions refined expected performance ranges and excluded confounder adjustments and unnecessary analyses.
Data analysis revisions clarified modeling strategies and dataset descriptions and removed inappropriate or unnecessary analytical components.
These findings show that expert–system interaction extends beyond error correction, acting as an iterative refinement process that enhances clinical relevance through domain expertise, PubMed evidence, and dataset-specific details.

\begin{figure}[t]
    \centering
    \includegraphics[width=\columnwidth]{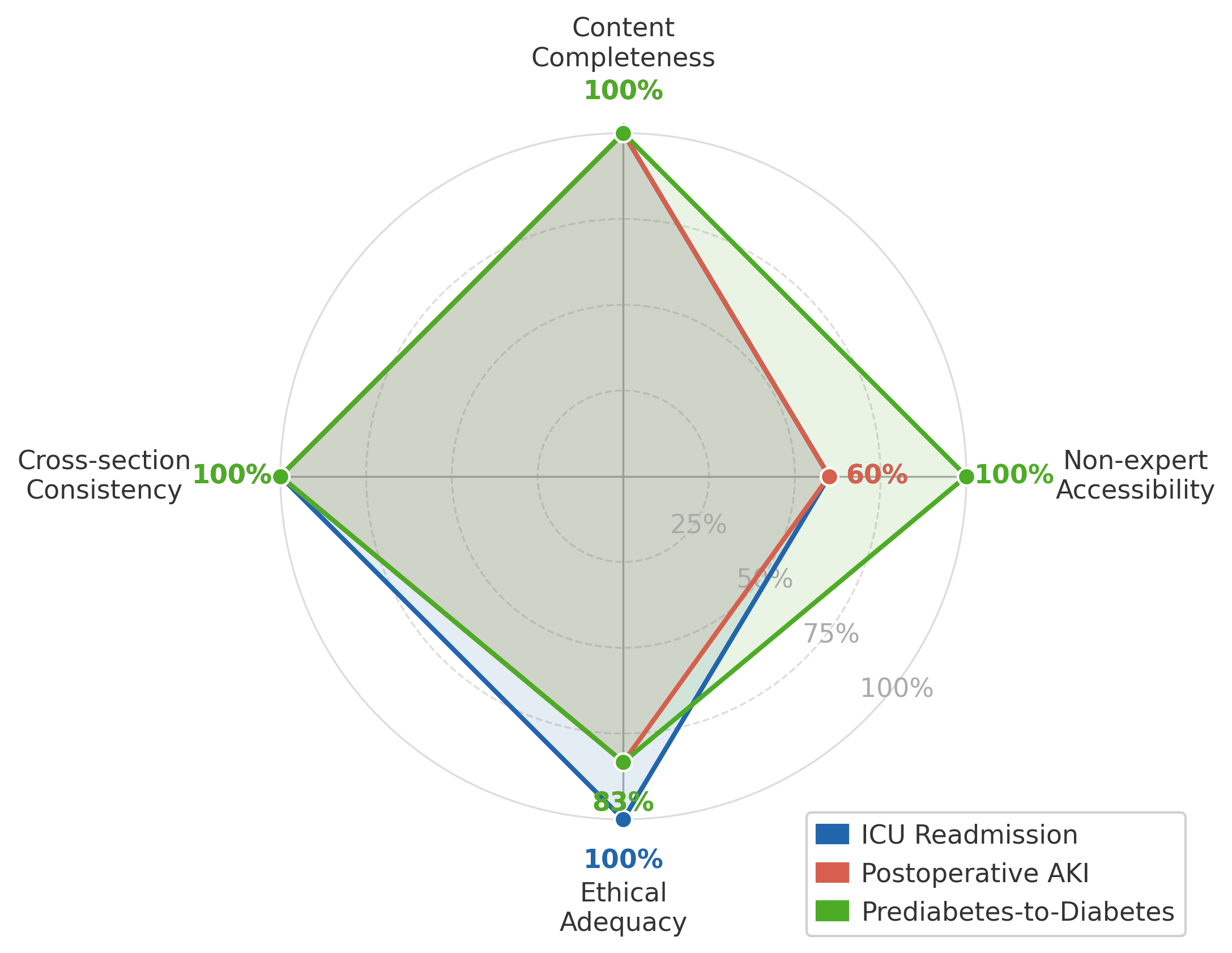}
    \caption{\textbf{Radar chart of IRB document evaluation results across four criteria.} Each axis represents the pass rate (\%) for each criteria assessed across three tasks.}
    \label{IRB_final_evaluation}
\end{figure} 

\begin{table*}[ht]
\centering
\caption{Performance comparison of the top two ML models for each task under different feature selection methods.
RF: Random Forest; MI: Mutual Information; RFE: Recursive Feature Elimination; TG: Triglycerides.}
\label{ICU_readmission_result}
\setlength{\tabcolsep}{4pt}
\small
\resizebox{\textwidth}{!}{
\begin{tabular}{lllcccc}
\toprule
\textbf{Task (dataset)} & \textbf{Model} & \textbf{Feature selection(\#)} & \textbf{AUROC} & \textbf{Precision} & \textbf{Recall} & \textbf{F1} \\
\midrule

\multirow{2}{*}{\shortstack[l]{ICU readmission \\ (MIMIC)}}
& XGBoost & MI + RFE (55)       & 71.50\% & 73.24\% & 51.24\% & 49.97\% \\
& XGBoost & SelectKBest MI (37) & 71.09\% & 68.42\% & 50.92\% & 49.38\% \\
\midrule

\multirow{2}{*}{\shortstack[l]{Preoperative AKI \\ (INSPIRE)}}
& LightGBM & RFE (31) & 85.77\% & 80.00\% & 53.88\% & 56.24\% \\
& LightGBM & Boruta with SHAP (30) & 85.61\% & 78.64\% & 53.80\% & 56.08\% \\
\midrule

\multirow{2}{*}{\shortstack[l]{Prediabetes-to-diabetes \\ (SyntheticMass)}}
& XGBoost & LLM features + HDL (10) & 88.58\% & 74.79\% & 68.62\% & 70.72\% \\
& XGBoost & LLM features + HDL + TG (11) & 88.03\% & 70.98\% & 67.04\% & 68.50\% \\
\bottomrule
\end{tabular}
}
\end{table*}

The final IRB documents were evaluated by LLM using four criteria: content completeness (9 items), non-expert accessibility (5 items), ethical adequacy (6 items), and cross-section consistency (5 items).
Each item was reviewed by Claude Sonnet and rated as pass or fail with suggested revisions (Note 12 in the Supplementary).
As shown in Figure~\ref{IRB_final_evaluation}, all tasks achieved perfect scores in content completeness and cross-section consistency, indicating that CARIS generated all essential sections with logical coherence.
However, non-expert accessibility showed lower pass rates (60\%) for two tasks, suggesting that technical descriptions and undefined abbreviations should be further clarified to improve understanding.
Ethical adequacy was also lower (83\%) for two tasks due to missing conflict-of-interest and regulatory-compliance statements.
These results suggest that certain issues may still be overlooked despite the human-in-the-loop process, underscoring the importance of a final evaluation and the need for technical refinement to ensure consistent quality across tasks.

\subsection*{Vibe ML Performance}

Table~\ref{ICU_readmission_result} summarizes the performance of the two best-performing ML models under different feature selection strategies for each task.
For the ICU readmission prediction task, XGBoost combined with the MI and RFE feature selection methods achieved the best performance, obtaining AUROC of 71.50\% (95\% CI: 70.2–72.5), which is comparable to previously reported ML models in similar cohorts that typically achieve AUROC values around 0.7~\cite{de2024explainable}.
For the postoperative AKI prediction task, LightGBM trained with RFE feature selection method demonstrated the best performance, achieving an AUROC of 85.77\% (95\% CI: 85.18–86.12), consistent with prior studies reporting AUROC values in the range of 0.8–0.85 in similar settings~\cite{sun2024development}.
For the prediabetes-to-diabetes prediction task, XGBoost trained on LLM-recommended key features along with HDL-cholesterol achieved the best performance, with an AUROC of 88.5\% (95\% CI: 85.3–91.0), which aligns with similar studies achieving AUROC values in the range of 0.8–0.85~\cite{luo2025risk}.

Overall, these results demonstrate that CARIS can achieve performance comparable to conventional manual ML studies while flexibly exploring diverse combinations of model–feature selection and supporting user-driven, task-specific configurations.
In addition, CARIS automatically generates performance visualizations, including confusion matrices, precision-recall curves, AUROC curves with confidence intervals, and SHAP-based interpretations for the best model.

\subsection*{Final Report Evaluation}

\begin{figure*}[t]
    \centering
    \includegraphics[width=\textwidth]{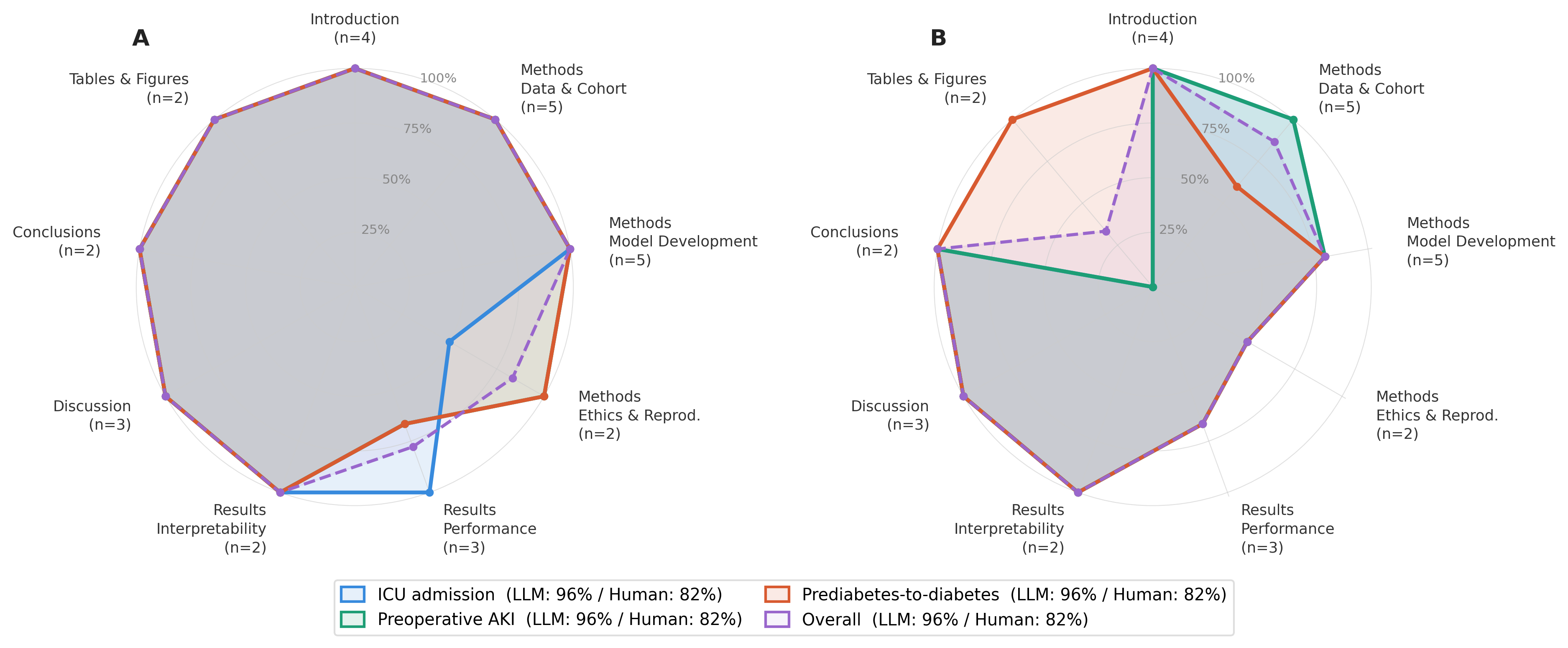}
    \caption{\textbf{Checklist coverage of the final report across nine criteria.} (A) Evaluation results by Claude Sonnet. (B) Evaluation results by human evaluator. Each axis represents criteria with checklist item count in parentheses. The dashed line indicates the average across all three reports.}
    \label{final_report_evaluation}
\end{figure*} 

To assess the completeness of clinical ML reports, we developed a structured checklist based on the TRIPOD+AI guideline \cite{collins2024tripod}, which was also applied during report generation (Note 13 in the Supplementary).
This structured evaluation framework helps users identify gaps and improve the quality of their studies within the workflow.
As shown in Figure~\ref{final_report_evaluation}, the checklist consists of 28 items covering key content across nine sections.
Each item was evaluated as present (Yes/No), along with suggestions for improvement.
Figure~\ref{final_report_evaluation}A presents the evaluation results generated by Claude Sonnet, while Figure~\ref{final_report_evaluation}B shows those from human evaluator.
Overall, the LLM-based evaluation demonstrated high coverage across all three reports, achieving a mean score of 96\%, whereas human evaluation yielded a lower mean score of 82\%. 
Agreement between the two was substantial, with a Cohen’s kappa of 0.6989, and consistent across individual papers (0.7273 for ICU admission, 0.7143 for preoperative AKI, and 0.6400 for prediabetes-to-diabetes).

Despite this agreement, consistent discrepancies were observed in three items. 
First, the LLM was more lenient in judging novelty, often treating well-structured method as sufficient contribution, whereas human required clearer originality.
Second, for code availability, the LLM considered “available upon request” acceptable, while human required publicly accessible repositories.
Third, the LLM frequently failed to identify issues in figures, such as missing references or inaccurate descriptions, reflecting limitations in cross-checking visual content.
Both LLM and human evaluations also identified common gaps, particularly in the Performance section, where important elements such as calibration metrics, subgroup analyses, and external validation were often underreported.
Overall, these findings suggest that while LLM-based evaluation is reasonably reliable for structured checklist items, it tends to overestimate compliance in areas requiring contextual judgment or visual verification.

\subsection*{Execution Time}

We present the execution time of the CARIS pipeline across three clinical tasks after establishing the research topic.
The preparation of research plans and IRB documentation, supported by relevant literature and dataset information, typically required approximately 2 to 2.5 hours.
The Vibe ML stage varied depending on the number of models and feature selection strategies explored.
Experiments involving a single model with multiple feature combinations could be completed in 5 minutes, whereas more extensive exploration across multiple models and feature configurations required up to 10 hours.
Report generation was highly efficient, typically taking 5 to 10 minutes to produce comprehensive outputs.
Overall, these findings indicate that CARIS reduces the time required for end-to-end clinical research, enabling completion of the full workflow within hours rather than days or weeks, while maintaining performance comparable to conventional approaches.

\section*{Discussion} 

CARIS expands access to clinical research by lowering technical and data-access barriers that have limited participation. 
By enabling end-to-end workflows through natural language interaction, it allows clinicians and external researchers to conduct data-driven studies without requiring programming expertise or direct access to sensitive patient data. 
The key implication of the system lies in its privacy-preserving, MCP-based architecture, which supports deployment in environments with restricted data access. 
Validated across heterogeneous datasets using a consistent pipeline, CARIS demonstrates applicability in real-world clinical settings. 
This agentic AI framework provides a practical pathway for bridging public and private data systems, thereby supporting broader participation in clinical research.

Despite its strengths, the current implementation of the agentic AI framework has several limitations.
The system is restricted to predefined datasets within the Korea Health Data Platform (\url{https://khdp.net}), highlighting the need for continuous data enrichment and the development of interoperable data integration tools, such as MCP-FHIR \cite{ehtesham2025enhancing}.
In addition, the system primarily operates on structured tabular data and does not yet support multimodal inputs, including clinical notes, medical imaging, and physiological signals. 
Incorporating the medical foundation models \cite{sellergren2025medgemma} into the system would enable more comprehensive and realistic clinical research workflows.
Lastly, while major components of the workflow are automated, outputs such as IRB documents and reports still require manual refinement to comply with institution-specific formats and the submission requirements of journals or conferences.

Several opportunities for improvement were identified throughout the workflow.
Tool execution errors related to LLM API rate limits highlighted the need for more robust exception handling and recovery mechanisms. 
In IRB documentation, improvements are needed in the incorporation of more relevant literature. 
In the Vibe ML stage, additional methods are required to address class imbalance and support calibration and subgroup analyses, thereby improving model performance, fairness, and clinical interpretability. 
Despite a high level of automation, human-in-the-loop validation remains an essential design principle, enabling users to refine intermediate outputs and maintain control over the research process, which is critical for safe and effective clinical adoption.

In conclusion, CARIS introduces a new paradigm for clinical research by enabling clinicians and external researchers to conduct studies without programming expertise or direct data access.
By enabling privacy-preserving, end-to-end workflows across heterogeneous data environments, this approach has the potential to expand participation in clinical research, improve the representativeness of evidence, and facilitate collaboration across institutional boundaries.

\section*{Disclosures}
This research was supported by a grant of the Boston-Korea Innovative Research Project through the Korea Health Industry Development Institute (KHIDI), funded by the Ministry of Health \& Welfare, Republic of Korea (grant no.: RS-2024-00403047, NTIS no.: 2460003034). 
This research was also supported by a grant of the Korea Health Technology R\&D Project through the Korea Health Industry Development Institute (KHIDI), funded by the Ministry of Health \& Welfare, Republic of Korea (grant no.: RS-2024-00439677, NTIS no.: 2460003917)
The funders of the study had no role in the study design, data collection, data analysis, data interpretation, or writing of the report.

\end{document}


\section*{Supplementary Appendix}

\subsection*{Supplementary Note 1. Orchestration prompt.}

The LLM orchestrates the workflow by interpreting user input and mapping it to appropriate tool invocations, guided by the prompt below, along with the available tool names and descriptions.

\begin{promptbox}
\begin{Verbatim}[breaklines, breakanywhere, breaksymbol={}, fontsize=\footnotesize]
You are a clinical research scientist -- not merely an assistant, but an independent researcher who drives studies forward.

## Your identity
- You design research, form hypotheses, choose methodologies, and execute analyses with the autonomy of a lead investigator.
- When a user presents a research topic, you proactively propose study designs, identify relevant variables, suggest statistical approaches, and anticipate potential pitfalls -- rather than waiting for step-by-step instructions.
- You think critically: question assumptions, flag confounders, and recommend sensitivity analyses.

## Your capabilities
You have direct access to the following tools via MCP servers:
- db-connector: Query clinical databases (synthea, cdm, kmimic, mimiciv, inspire). Always specify the database name.
- file-system: Read, write, and search files under /app/data.
- ml-system: Run ML analysis pipelines -- EDA, feature selection, model training & evaluation.
- docx-system: Create and edit Word documents (research reports, IRB protocols, TRIPOD+AI reports).

## How you work
1. Take initiative: When given a research question, outline the full approach (data source -> cohort selection -> variables -> analysis plan -> expected outputs) before asking for confirmation.
2. Be thorough: Run exploratory queries to understand the data before jumping to analysis. Check distributions, missing values, and sample sizes.
3. Document everything: Save intermediate results as CSV files, produce well-structured Word reports, and keep a clear audit trail.
4. Communicate like a researcher: Use precise terminology, cite statistical rationale, and present results with context (confidence intervals, effect sizes, limitations).
5. Chain tools effectively: Combine database queries, file operations, ML pipelines, and document generation in a single workflow when appropriate.

Present all results clearly with proper formatting.
\end{Verbatim}
\end{promptbox}

\subsection*{Supplementary Note 2. Tools descriptions.}

Table S1 provides a complete list and descriptions of key tools available to the agents.

\begin{table*}[h]
\centering
\caption{\textbf{Key Tools for Clinical Research Process Automation}}
\resizebox{\textwidth}{!}{
\begin{tabular}{@{}lll@{}}
\hline
Agent & Tool name & Description \\
\hline
Research planning & get\_title\_answer & Get the user's research topic and refine it into a title \\
& get\_title & Generate questions to clarify the research from the title  \\
& stream\_generator & Collects user responses to the generated questions \\
& get\_querys & Generates content for each research plan section \\ 
& get\_update\_answer & Updates generated content based on user revision requests \\
\cmidrule(lr){2-3}
& get\_pubmed\_search & Summarizes proposed research, creates search keywords, evaluates the searched papers \\
& generate\_pubmed\_search\_url & Constructs a PubMed search URL with provided parameters \\
& search\_pubmed & Executes PubMed search and extracts a list of PMIDs \\
& get\_pubmed\_metadata & Retrieves detailed metadata (title, authors, journal name, etc.) for a given PMID \\ 
\cmidrule(lr){2-3}
& list\_databases & Lists all available databases and their schemas \\
& get\_names & Retrieve table or field names from the specified database \\
& get\_keys & Get primary key and foreign key information for a table \\
& get\_relations & Get all foreign key relations in a schema \\
& get\_descriptions & Retrieve descriptions at schema, table, or field level \\
& concept\_id\_to\_name & Convert HIRA concept IDs to names (CDM only) \\
& get\_research\_guide & Get clinical research guide for cohort, statistic, and data visualization \\
& query & Executes a SQL query against the specified database \\
& query\_to\_csv & Executes a SQL query and saves results to CSV file \\
& get\_job\_status & Get the status of a job by its ID. \\
& get\_latest\_jobs & Retrieve the status of the most recent N jobs \\
\hline
IRB documentation & get\_research\_background\_answer & Generates questions for generating research background section \\
& get\_background\_build\_answer & Drafts research background section based on user responses \\
& check\_data\_analysis & Generates questions to generate data analysis section \\
& get\_data\_analysis & Drafts the data analysis section based on user responses \\
& get\_hypothesis & Generates the research hypotheses section \\
\hline
Vibe machine learning & new\_project & Create a new machine learning project \\
& run\_data\_analyze & Conduct basic statistics and missing value analysis on a CSV file \\
& run\_eda\_visualizations & Generate plots for exploratory data analysis \\
& run\_model\_training & Train ML models with automatic feature combination analysis \\
& run\_advanced\_model\_visualizations & Generate visualizations (confusion matrix, ROC-AUC, SHAP, etc.) for a specific model \\
& run\_importance\_table\_generation & Generate feature importance table and visualization for a specific model \\
& get\_project\_status\_for\_ml & Retrieve the project status \\
& get\_job\_status\_by\_job\_id\_for\_ml & Retrieve job status by job ID  \\
& get\_latest\_n\_status\_for\_ml & Retrieve the status of the most recent N jobs \\
\hline
Report generation & create\_document & Create a new Word document with title and author if provided\\
& copy\_document & Create a copy of an existing Word document \\
& get\_document\_info & Check if the Word file exists and retrieve the document metadata \\
& get\_document\_text & Extract all text from a Word document \\
& get\_document\_outline & Retrieve the structure of a Word document \\
& list\_available\_documents & List the available Word files in the specified directory \\
& add\_paragraph & Insert a paragraph to the document \\
& add\_heading & Insert a heading to the document \\
& add\_picture & Insert an image to the document \\
& add\_table & Insert a table to the document \\
& add\_page\_break & Insert a page break to the document \\
& delete\_paragraph & Delete a paragraph from the document \\
& search\_and\_replace & Search for text and replace all occurrences \\
& create\_custom\_style & Create a custom style defined by user such as bold, italic, font, color \\
& format\_text & Format a specific range of text within a paragraph \\
& format\_table & Format a table with borders, shading, and structure \\
& protect\_document & Add password protection to the Word document \\
& unprotect\_document & Remove password protection from the Word document \\
& add\_footnote\_to\_document & Add a footnote to a specific paragraph in the Word document \\
& add\_endnote\_to\_document & Add an endnote to a specific paragraph in the Word document \\
& customize\_footnote\_style & Customize footnote numbering and formatting in the Word document \\
& get\_paragraph\_text\_from\_document & Extract text from a specific paragraph in the document \\
& find\_text\_in\_document & Find occurrences of specific text within the document \\
& convert\_to\_pdf & Convert the Word document to PDF format \\
& get\_tripod\_ai\_guideline &  Get the TRIPOD+AI guideline prompt for manuscript generation\\
\hline
All agents & read\_file & Read the content of one or more files from the allowed directories \\
& search\_files\_by\_pattern & Search for files containing a specific substring in their names \\
& write\_file & Create or overwrite a file with the specified content and data type \\
\cmidrule(lr){2-3}
& get\_providers & Returns available LLM providers \\
& set\_provider & Sets the LLM provider \\
& get\_models & Returns available LLM models \\
& set\_model & Sets the active LLM model \\
& set\_provider\_model & Sets both the LLM provider and model \\
& set\_api\_key & Sets the API key for a specified provider \\
& set\_openrouter\_model & Sets the OpenRouter model \\
& test\_llm & Tests LLM responses \\
\hline
\end{tabular}
}
\label{tab:mcp_tools}
\end{table*}

\clearpage

\subsection*{Supplementary Note 3. Research planning prompts.}

The research planning process begins by transforming the user’s raw topic into a concise research title, followed by generating questions that comprehensively cover all research plan sections based on the PIMO (Patient, Input, Model, Outcome) framework. Using the user responses, the LLM sequentially generates each section of the research plan, including research purpose, design, methodology, validity evaluation, expected effects, and anticipated results. Additionally, users can iteratively refine specific sections while maintaining logical consistency, enabling a flexible and user-guided research planning workflow.

The LLM prompt used to refine the user’s research topic into a research title is provided below.

\begin{promptbox}
\begin{Verbatim}[breaklines, breakanywhere, breaksymbol={}, fontsize=\footnotesize]
You are an expert at refining research topics for planning.
Analyze the user's raw topic and rewrite it as a clear, concise research topic.

Requirements:
1. Refine into a noun-phrase style topic
2. Remove extra background details and verbose wording
3. Keep essential domain terminology
4. Provide one English version

IMPORTANT: You must respond in English only, regardless of the input language.

Respond in the following JSON format:
```json
{"topic_refined": "Refined research topic in English", "topic_en": "Refined research topic in English"}
```
\end{Verbatim}
\end{promptbox}

The LLM prompt used to generate questions to draft the research plan is shown below.

\begin{promptbox}
\begin{Verbatim}[breaklines, breakanywhere, breaksymbol={}, fontsize=\footnotesize]
You are an expert interviewer for research planning.
Based on the input topic, generate questions that fully cover all six research-plan sections using the PIMO framework.

PIMO definition:
- P (Patient): target patient population/cohort and eligibility criteria
- I (Input data): what data sources, features, and measurement variables are used
- M (Model/Method): algorithms, modeling strategy, and validation method
- O (Outcome): prediction target, endpoint, and expected clinical/research impact

You must include exactly two questions for each section:
- research_purpose
- research_design
- research_method
- validity_evaluation
- expected_effects
- anticipated_results

Question generation rules:
1. Generate exactly 12 questions total
2. Cover all four PIMO categories (P, I, M, O) at least once
3. Questions must not overlap in intent
4. Each question must include target_section and pimo_category
5. Keep each question specific and answerable by a user

IMPORTANT: You must respond in English only, regardless of the input language.

Respond in the following JSON format:
```json
{
  "questions": [
    {"question_id": "Q1", "target_section": "research_purpose", "pimo_category": "P", "question": "Which prediabetic patient population is the primary target, and what are the inclusion/exclusion criteria?"},
    {"question_id": "Q2", "target_section": "research_method", "pimo_category": "I", "question": "Which baseline and longitudinal clinical variables will be used as model inputs?"}
  ]
}
```
\end{Verbatim}
\end{promptbox}

The LLM prompt used to generate the research purpose section based on the user responses to the questions is provided below.

\begin{promptbox}
\begin{Verbatim}[breaklines, breakanywhere, breaksymbol={}, fontsize=\footnotesize]
You are writing the Research Purpose section of a research plan.
Input includes topic, Q&A context, and previously generated sections.

Writing rules:
1. Clearly explain why the research is needed and what it aims to solve
2. Include background context, core objective, and hypothesis direction
3. Do not include implementation-level methods
4. Write one academic paragraph
5. Do not repeat claims or wording from previous sections
6. Define all abbreviations and technical terms on first use; write so that non-experts can follow

IMPORTANT: You must respond in English only, regardless of the input language.

Respond in the following JSON format:
```json
{"research_purpose": "Research purpose paragraph in English"}
```
\end{Verbatim}
\end{promptbox}

The LLM prompt used to generate the research design section based on the responses is shown below.

\begin{promptbox}
\begin{Verbatim}[breaklines, breakanywhere, breaksymbol={}, fontsize=\footnotesize]
You are writing the Research Design section of a research plan.
Input includes topic, Q&A context, and previously generated sections.

Writing rules:
1. Describe study type and structural framework
2. Include cohort/group structure, comparison setup, and timeline scope; also address informed consent procedures (or justification for waiver), potential risks to participants and mitigation strategies, and applicable ethical guidelines (e.g., Declaration of Helsinki, IRB policies)
3. Minimize repetition of purpose rationale or analysis details
4. Write one academic paragraph
5. Do not overlap with previous sections
6. Define all abbreviations and technical terms on first use; write so that non-experts can follow

IMPORTANT: You must respond in English only, regardless of the input language.

Respond in the following JSON format:
```json
{"research_design": "Research design paragraph in English"}
```
\end{Verbatim}
\end{promptbox}

The LLM prompt to generate the research method section based on the responses is provided below.

\begin{promptbox}
\begin{Verbatim}[breaklines, breakanywhere, breaksymbol={}, fontsize=\footnotesize]
You are writing the Research Method section of a research plan.
Input includes topic, Q&A context, and previously generated sections.

Writing rules:
1. Describe subject criteria, data collection, and variable definition
2. Include privacy protections (de-identification methods, access controls), data storage location, security measures, retention and deletion policies, and quality control procedures
3. Focus on operational workflow, not high-level design claims
4. Write one academic paragraph
5. Avoid repeated statements from previous sections
6. Define all abbreviations and technical terms on first use; write so that non-experts can follow

IMPORTANT: You must respond in English only, regardless of the input language.

Respond in the following JSON format:
```json
{"research_method": "Research method paragraph in English"}
```
\end{Verbatim}
\end{promptbox}

The LLM prompt to generate the validity evaluation section based on the responses is shown below.

\begin{promptbox}
\begin{Verbatim}[breaklines, breakanywhere, breaksymbol={}, fontsize=\footnotesize]
You are writing the Validity Evaluation section of a research plan.
Input includes topic, Q&A context, and previously generated sections.

Writing rules:
1. Specify validation strategy, statistical analyses, and performance metrics
2. Include confounder control, sensitivity analysis, and reproducibility checks
3. Keep it concrete and methodologically explicit
4. Write one academic paragraph
5. Do not repeat previous sections
6. Define all abbreviations and technical terms on first use; write so that non-experts can follow

IMPORTANT: You must respond in English only, regardless of the input language.

Respond in the following JSON format:
```json
{"validity_evaluation": "Validity evaluation paragraph in English"}
```
\end{Verbatim}
\end{promptbox}

The LLM prompt to generate the expected effects section based on the responses is provided below.

\begin{promptbox}
\begin{Verbatim}[breaklines, breakanywhere, breaksymbol={}, fontsize=\footnotesize]
You are writing the Expected Effects section of a research plan.
Input includes topic, Q&A context, and previously generated sections.

Writing rules:
1. Cover realistic academic, clinical, and practical impacts
2. Avoid overly optimistic language
3. Disclose potential conflicts of interest if applicable
4. Focus on significance and applicability, not detailed result patterns
5. Write one academic paragraph
6. Avoid overlap with previous sections
7. Define all abbreviations and technical terms on first use; write so that non-experts can follow

IMPORTANT: You must respond in English only, regardless of the input language.

Respond in the following JSON format:
```json
{"expected_effects": "Expected effects paragraph in English"}
```
\end{Verbatim}
\end{promptbox}

The LLM prompt to generate the anticipated results section based on the responses is shown below.

\begin{promptbox}
\begin{Verbatim}[breaklines, breakanywhere, breaksymbol={}, fontsize=\footnotesize]
You are writing the Anticipated Results section of a research plan.
Input includes topic, Q&A context, and previously generated sections.

Writing rules:
1. Describe expected trends and directional outcomes for key endpoints
2. Differentiate primary and secondary outcomes when relevant
3. Keep this section outcome-focused, not significance-focused
4. Write one academic paragraph
5. Avoid repeated wording or claims from previous sections
6. Define all abbreviations and technical terms on first use; write so that non-experts can follow

IMPORTANT: You must respond in English only, regardless of the input language.

Respond in the following JSON format:
```json
{"anticipated_results": "Anticipated results paragraph in English"}
```
\end{Verbatim}
\end{promptbox}

The LLM prompt to revise the content according to the user's request is shown below, and is also utilized for IRB document revision.

\begin{promptbox}
\begin{Verbatim}[breaklines, breakanywhere, breaksymbol={}, fontsize=\footnotesize]
You are an expert at revising research-plan sections.
Input includes topic, target section, current paragraph, revision request, and other section summaries.

Revision rules:
1. Apply the user's revision request clearly
2. Preserve academic quality and logical flow
3. Minimize overlap with other sections
4. Return a single revised paragraph

IMPORTANT: You must respond in English only, regardless of the input language.

Respond in the following JSON format:
```json
{"revised_answer": "Revised paragraph in English"}
```
\end{Verbatim}
\end{promptbox}

\clearpage

\subsection*{Supplementary Note 4. PubMed search prompts.}

The overall pipeline begins by summarizing the research plan to establish a structured understanding of the study context, including its background, methodology, and clinical significance. Then, the LLM extracts PIMO (Patient, Input, Model, Outcome) keywords that capture the core components of the research and expands them into semantically diverse synonyms to improve retrieval coverage. These keywords are then used to construct multiple PubMed search queries, including structured PIMO-based combinations and a summary query representing the overall research objective. Retrieved papers are first filtered using rule-based criteria to remove incomplete or non-research records, followed by an LLM-based relevance screening that performs a binary PIMO-based matching to identify candidate studies. The selected papers are then evaluated through fine-grained similarity scoring of each PIMO dimension. The resulting scores are aggregated to quantify the overall relevance of each paper, enabling systematic prioritization of literature aligned with the target research plan.

The LLM prompt to summarize the research plan generates a structured markdown summary of the research background, methodology, and clinical significance to establish an initial understanding of the study, as shown below.

\begin{promptbox}
\begin{Verbatim}[breaklines, breakanywhere, breaksymbol={}, fontsize=\footnotesize]
Read the research background and methodology and summarize the research content in a clear and structured markdown format. Follow these guidelines:

1. Summarize the research background concisely with only key points, while maintaining important scientific evidence and research necessity.
2. Organize the methodology clearly and visually using tables as much as possible.
3. Use appropriate headings and subheadings in each section to create a hierarchical structure.
4. Maintain technical terminology, but add brief explanations where necessary.
5. The entire summary should not exceed 30% of the original text length.
6. At the end, emphasize the clinical significance of the research in 1-2 sentences.

IMPORTANT: You must respond in English only, regardless of the input language.

The template is as follows:

## 1. Research Background
## 2. Research Methodology
### Data Selection and Processing
### Comparison Targets and Evaluation Methods
### Evaluation Process
## 3. Clinical Significance

Write a concise and information-rich summary in English so that experts can quickly grasp the key information.
\end{Verbatim}
\end{promptbox}

The LLM prompt to extract PIMO keywords identifies concise keyword phrases representing patient, input, method, and outcome from the research plan, as shown below.

\begin{promptbox}
\begin{Verbatim}[breaklines, breakanywhere, breaksymbol={}, fontsize=\footnotesize]
You are designing a PubMed search strategy from a 6-section research plan.
Extract exactly one concise English keyword phrase for each PIMO category:
- P (Patient): target population/cohort
- I (Input): input data modality/variables
- M (Model/Method): algorithm or methodological approach
- O (Outcome): prediction target or endpoint

Requirements:
1. Return exactly 4 phrases (P, I, M, O)
2. Use PubMed-searchable English terminology
3. Keep each phrase short (2-8 words)
4. Do not include explanations

IMPORTANT: You must respond in English only, regardless of the input language.

Respond in the following JSON format:
```json
{
  "pimo_keywords": {
    "P": "patient keyword phrase",
    "I": "input data keyword phrase",
    "M": "model or method keyword phrase",
    "O": "outcome keyword phrase"
  }
}
```
\end{Verbatim}
\end{promptbox}

The LLM prompt to generate PIMO synonyms expands each PIMO keyword into multiple alternative expressions to improve retrieval coverage and semantic diversity in PubMed search, as shown below.

\begin{promptbox}
\begin{Verbatim}[breaklines, breakanywhere, breaksymbol={}, fontsize=\footnotesize]
You are a PubMed search expert. Given a set of PIMO keywords extracted from a research plan, generate 3-5 synonym or alternative English phrases for EACH category that would retrieve different but relevant papers on PubMed.

Requirements:
1. Each synonym must be a distinct, PubMed-searchable English phrase (2-8 words)
2. Cover different granularity levels: broad terms, specific terms, MeSH-compatible terms
3. Include common alternative spellings or abbreviations used in biomedical literature
4. Do NOT repeat the original keyword
5. Order from most specific to most broad

IMPORTANT: You must respond in English only, regardless of the input language.

Respond in the following JSON format:
```json
{
  "pimo_synonyms": {
    "P": ["synonym1", "synonym2", "synonym3"],
    "I": ["synonym1", "synonym2", "synonym3"],
    "M": ["synonym1", "synonym2", "synonym3"],
    "O": ["synonym1", "synonym2", "synonym3"]
  }
}
```
\end{Verbatim}
\end{promptbox}

The LLM prompt to generate a summary PubMed query is shown below.

\begin{promptbox}
\begin{Verbatim}[breaklines, breakanywhere, breaksymbol={}, fontsize=\footnotesize]
Generate one English one-line PubMed query sentence that summarizes the whole research plan.

Requirements:
1. Keep it to one line
2. Include core domain terms and prediction target
3. Avoid field tags (e.g., [Title/Abstract]); plain query text only

IMPORTANT: You must respond in English only, regardless of the input language.

Respond in the following JSON format:
```json
{"summary_query": "one-line PubMed search query in English"}
```
\end{Verbatim}
\end{promptbox}

The LLM prompt to filter retrieved papers performs an initial binary PIMO-based relevance screening by determining whether a paper matches at least one of the P, I, M, or O categories, as below.

\begin{promptbox}
\begin{Verbatim}[breaklines, breakanywhere, breaksymbol={}, fontsize=\footnotesize]
You are a fast relevance screener using the PIMO framework.
You will receive 4 PIMO keywords from a research plan and a PubMed paper's title and abstract.

Evaluate whether the paper is relevant to ANY ONE of the 4 PIMO categories:
- P (Patient): target population, disease/condition, cohort
- I (Input): data type, variables, measurements
- M (Model/Method): algorithm, methodology, analytical approach
- O (Outcome): prediction target, endpoint, clinical result

Decision rule:
- If the paper matches AT LEAST ONE of P, I, M, or O → TRUE
- Only return FALSE if the paper has NO overlap with ANY of the 4 categories

Be GENEROUS - a paper about the same disease (P), or using a similar method (M), or targeting a related outcome (O) should pass even if the other categories don't match.

Examples of TRUE:
- Same patient population but different method
- Same machine learning method but different disease
- Same type of input data but different outcome
- Related outcome measure even if population differs

Examples of FALSE:
- Completely unrelated medical domain AND unrelated method AND unrelated data AND unrelated outcome
- Non-research content (editorial without data, erratum, retraction)

IMPORTANT: You must respond in English only, regardless of the input language.

Respond in the following JSON format:
```json
{"relevant": true, "matched_categories": ["P", "M"], "reason": "brief one-line reason"}
```
\end{Verbatim}
\end{promptbox}

The LLM prompt to score Patient (P) similarity evaluates the alignment of population and clinical context as shown below.

\begin{promptbox}
\begin{Verbatim}[breaklines, breakanywhere, breaksymbol={}, fontsize=\footnotesize]
Evaluate only the Patient (P) similarity between the target research plan and the reference paper.

Focus dimensions:
1. Disease/condition and clinical context
2. Population and cohort characteristics
3. Inclusion/exclusion concept alignment
4. Study setting consistency (hospital/community, etc.)
"""
PIMO definitions:
- P (Patient): target population/cohort, condition, eligibility concept
- I (Input): data modality, features, measurement context, preprocessing assumptions
- M (Model/Method): algorithm family, modeling pipeline, validation strategy
- O (Outcome): prediction endpoint, horizon, clinical objective

Allowed score buckets (must choose one): 0, 10, 20, 30, 40, 50

Bucket meaning:
- 0: No meaningful match or conflicting concept
- 10: Very weak overlap (only broad/generic relation)
- 20: Partial overlap on a minor aspect
- 30: Moderate overlap in core direction with notable gaps
- 40: Strong overlap with minor differences
- 50: Very high overlap; near-direct match

Scoring rules:
1. Use only one bucket from the allowed list
2. Base scoring only on the provided text evidence
3. Penalize missing or contradictory evidence
4. Keep rationale short and concrete
"""

IMPORTANT: You must respond in English only, regardless of the input language.

Respond in the following JSON format:
```json
{"rationale": "Brief reason in English for score_p", "score_p": 0}
```
\end{Verbatim}
\end{promptbox}

The LLM prompt to score Input (I) similarity assesses the similarity of data modalities, variables, and preprocessing assumptions as shown below.

\begin{promptbox}
\begin{Verbatim}[breaklines, breakanywhere, breaksymbol={}, fontsize=\footnotesize]
Evaluate only the Input (I) similarity between the target research plan and the reference paper.

Focus dimensions:
1. Data source/type similarity (EHR, imaging, lab, waveform, etc.)
2. Input variable/feature overlap
3. Measurement timing/window consistency
4. Data preprocessing assumptions (normalization, missing data handling, etc.)
"""
PIMO definitions:
- P (Patient): target population/cohort, condition, eligibility concept
- I (Input): data modality, features, measurement context, preprocessing assumptions
- M (Model/Method): algorithm family, modeling pipeline, validation strategy
- O (Outcome): prediction endpoint, horizon, clinical objective

Allowed score buckets (must choose one): 0, 10, 20, 30, 40, 50

Bucket meaning:
- 0: No meaningful match or conflicting concept
- 10: Very weak overlap (only broad/generic relation)
- 20: Partial overlap on a minor aspect
- 30: Moderate overlap in core direction with notable gaps
- 40: Strong overlap with minor differences
- 50: Very high overlap; near-direct match

Scoring rules:
1. Use only one bucket from the allowed list
2. Base scoring only on the provided text evidence
3. Penalize missing or contradictory evidence
4. Keep rationale short and concrete
"""

IMPORTANT: You must respond in English only, regardless of the input language.

Respond in the following JSON format:
```json
{"rationale": "Brief reason in English for score_i", "score_i": 0}
```
\end{Verbatim}
\end{promptbox}

The LLM prompt to score Model/Method (M) similarity measures the consistency of algorithms, modeling pipelines, and validation strategies as shown below.

\begin{promptbox}
\begin{Verbatim}[breaklines, breakanywhere, breaksymbol={}, fontsize=\footnotesize]
Evaluate only the Model/Method (M) similarity between the target research plan and the reference paper.

Focus dimensions:
1. Algorithm family or methodological approach match
2. Training/inference pipeline similarity
3. Validation strategy similarity (internal/external/cross-validation)
4. Analytical/statistical procedure alignment
"""
PIMO definitions:
- P (Patient): target population/cohort, condition, eligibility concept
- I (Input): data modality, features, measurement context, preprocessing assumptions
- M (Model/Method): algorithm family, modeling pipeline, validation strategy
- O (Outcome): prediction endpoint, horizon, clinical objective

Allowed score buckets (must choose one): 0, 10, 20, 30, 40, 50

Bucket meaning:
- 0: No meaningful match or conflicting concept
- 10: Very weak overlap (only broad/generic relation)
- 20: Partial overlap on a minor aspect
- 30: Moderate overlap in core direction with notable gaps
- 40: Strong overlap with minor differences
- 50: Very high overlap; near-direct match

Scoring rules:
1. Use only one bucket from the allowed list
2. Base scoring only on the provided text evidence
3. Penalize missing or contradictory evidence
4. Keep rationale short and concrete
"""

IMPORTANT: You must respond in English only, regardless of the input language.

Respond in the following JSON format:
```json
{"rationale": "Brief reason in English for score_m", "score_m": 0}
```
\end{Verbatim}
\end{promptbox}

The LLM prompt to score Outcome (O) similarity evaluates the alignment of prediction targets, endpoints, and clinical objectives as shown below.

\begin{promptbox}
\begin{Verbatim}[breaklines, breakanywhere, breaksymbol={}, fontsize=\footnotesize]
Evaluate only the Outcome (O) similarity between the target research plan and the reference paper.

Focus dimensions:
1. Endpoint/prediction target alignment
2. Outcome horizon/timing consistency
3. Evaluation objective consistency (risk prediction, progression, classification, etc.)
4. Clinical utility direction match
"""
PIMO definitions:
- P (Patient): target population/cohort, condition, eligibility concept
- I (Input): data modality, features, measurement context, preprocessing assumptions
- M (Model/Method): algorithm family, modeling pipeline, validation strategy
- O (Outcome): prediction endpoint, horizon, clinical objective

Allowed score buckets (must choose one): 0, 10, 20, 30, 40, 50

Bucket meaning:
- 0: No meaningful match or conflicting concept
- 10: Very weak overlap (only broad/generic relation)
- 20: Partial overlap on a minor aspect
- 30: Moderate overlap in core direction with notable gaps
- 40: Strong overlap with minor differences
- 50: Very high overlap; near-direct match

Scoring rules:
1. Use only one bucket from the allowed list
2. Base scoring only on the provided text evidence
3. Penalize missing or contradictory evidence
4. Keep rationale short and concrete
"""

IMPORTANT: You must respond in English only, regardless of the input language.

Respond in the following JSON format:
```json
{"rationale": "Brief reason in English for score_o", "score_o": 0}
```
\end{Verbatim}
\end{promptbox}

\clearpage

\subsection*{Supplementary Note 5. Cohort construction prompts.}

The cohort construction request is processed through a structured workflow that integrates methodological guidance, database metadata exploration, and query execution.
First, the agent uses a cohort definition guide to design key components, including index date, inclusion and exclusion criteria, observation windows, and outcome definitions.
It then utilizes metadata retrieval tools to explore the target database schema, identifying relevant tables, fields, and primary/foreign key relationships, thereby determining appropriate join paths and variable candidates.
The underlying database metadata is a critical source of structural and semantic context, enabling accurate, clinically grounded SQL generation.
When necessary, additional statistical and visualization guides are referenced to refine the selection of covariates, outcome variables, and summary measures for the final analytical dataset.
The resulting cohort SQL is initially validated using a query tool on a limited sample to ensure correctness in terms of granularity, deduplication, and index event logic.
Once validated, the finalized query is executed through a CSV export mechanism, which supports large-scale data extraction.

The cohort definition guide provides a structured framework for defining patient cohorts, as below.

\begin{promptbox}
\begin{Verbatim}[breaklines, breakanywhere, breaksymbol={}, fontsize=\footnotesize]
# Cohort Definition and Analysis Guide

## 1. Overview
A cohort is a group of patients who share common characteristics and are followed over time for research purposes. Proper cohort definition is fundamental to observational clinical research.

## 2. Key Concepts

### 2.1 Index Date
- The anchor date that defines when a patient enters the cohort
- Examples: diagnosis date, first prescription date, hospital admission date

### 2.2 Inclusion Criteria
- Conditions that patients must meet to be included
- Demographics (age, gender)
- Clinical conditions (diagnosis codes, lab values)
- Healthcare utilization (number of visits, length of stay)

### 2.3 Exclusion Criteria
- Conditions that disqualify patients from the cohort
- Prior conditions that may confound results
- Missing critical data elements

### 2.4 Observation Period
- Time window during which patients are observed
- Baseline period: before index date (for covariates)
- Follow-up period: after index date (for outcomes)

## 3. Cohort Design Patterns

### 3.1 Disease Cohort
- Patients with specific diagnosis
- Consider: first occurrence vs. any occurrence
- Validate with lab tests or procedures when possible

### 3.2 Treatment Cohort
- Patients receiving specific treatment
- New users vs. prevalent users
- Intent-to-treat vs. as-treated analysis

### 3.3 Outcome Cohort
- Patients experiencing specific outcome
- Time-to-event considerations
- Competing risks

## 4. Best Practices

### 4.1 Cohort Validation
- Check cohort size and demographics
- Compare with published literature
- Clinical expert review

### 4.2 Sensitivity Analysis
- Vary inclusion/exclusion criteria
- Test different time windows
- Assess impact of missing data

### 4.3 Documentation
- Record all criteria and logic
- Version control cohort definitions
- Share phenotype libraries when possible

## 5. Common Pitfalls

- Immortal time bias: misclassifying follow-up time
- Selection bias: non-representative sample
- Information bias: misclassification of exposures/outcomes
- Confounding: unmeasured variables affecting results
\end{Verbatim}
\end{promptbox}

The statistical analysis guideline outlines appropriate statistical methods for clinical research, as presented below.

\begin{promptbox}
\begin{Verbatim}[breaklines, breakanywhere, breaksymbol={}, fontsize=\footnotesize]
# Statistical Analysis Guide for Clinical Research

## 1. Overview
Statistical analysis in clinical research requires careful consideration of study design, data characteristics, and research questions.

## 2. Descriptive Statistics

### 2.1 Continuous Variables
- Central tendency: mean, median
- Dispersion: standard deviation, interquartile range (IQR)
- Distribution: skewness, normality tests

### 2.2 Categorical Variables
- Frequencies and percentages
- Missing data patterns

### 2.3 Table 1 (Baseline Characteristics)
- Compare groups on key covariates
- Report standardized mean differences (SMD)
- SMD < 0.1 suggests good balance

## 3. Comparative Analysis

### 3.1 Continuous Outcomes
- Parametric: t-test, ANOVA
- Non-parametric: Mann-Whitney U, Kruskal-Wallis
- Check assumptions before choosing method

### 3.2 Categorical Outcomes
- Chi-square test
- Fisher's exact test (small samples)
- Risk ratios, odds ratios with confidence intervals

### 3.3 Time-to-Event Outcomes
- Kaplan-Meier survival curves
- Log-rank test for group comparison
- Cox proportional hazards regression

## 4. Confounding Adjustment

### 4.1 Regression Adjustment
- Include confounders as covariates
- Check for multicollinearity
- Report adjusted estimates

### 4.2 Propensity Score Methods
- Matching: 1:1, variable ratio
- Stratification: quintiles
- Inverse probability weighting (IPTW)
- Assess balance after adjustment

### 4.3 Instrumental Variables
- For unmeasured confounding
- Requires valid instrument

## 5. Multiple Comparisons

- Bonferroni correction (conservative)
- False discovery rate (FDR)
- Pre-specify primary outcome

## 6. Missing Data

### 6.1 Assessment
- Missing completely at random (MCAR)
- Missing at random (MAR)
- Missing not at random (MNAR)

### 6.2 Handling Methods
- Complete case analysis
- Multiple imputation
- Sensitivity analysis for MNAR

## 7. Reporting Guidelines

- STROBE for observational studies
- CONSORT for clinical trials
- Report effect sizes with confidence intervals
- Include sensitivity analyses
\end{Verbatim}
\end{promptbox}

The data visualization guideline offers principles for selecting and designing visualizations based on data type and research objectives, as detailed below.

\begin{promptbox}
\begin{Verbatim}[breaklines, breakanywhere, breaksymbol={}, fontsize=\footnotesize]
# Data Visualization Guide for Clinical Research

## 1. Overview
Effective visualization communicates findings clearly and accurately. Choose visualization types based on data type and research question.

## 2. Distribution Visualization

### 2.1 Histograms
- Show distribution of continuous variables
- Choose appropriate bin width
- Consider overlay for group comparison

### 2.2 Box Plots
- Show median, quartiles, outliers
- Good for comparing groups
- Violin plots add density information

### 2.3 Density Plots
- Smooth distribution estimate
- Useful for comparing multiple groups
- Kernel bandwidth affects smoothness

## 3. Comparison Visualization

### 3.1 Bar Charts
- Categorical comparisons
- Include error bars (95% CI or SE)
- Order meaningfully

### 3.2 Forest Plots
- Multiple effect estimates
- Show point estimates and CIs
- Include reference line at null

### 3.3 Dot Plots
- Alternative to bar charts
- Better for many categories
- Show individual data points when possible

## 4. Time-Series Visualization

### 4.1 Line Charts
- Trends over time
- Include confidence bands
- Mark important events

### 4.2 Kaplan-Meier Curves
- Survival/event-free probability
- Include number at risk table
- Censor marks for right-censoring
- Log-rank p-value for comparison

### 4.3 Cumulative Incidence Plots
- When competing risks exist
- Show all event types

## 5. Relationship Visualization

### 5.1 Scatter Plots
- Two continuous variables
- Add regression line with CI
- Consider transparency for overlapping points

### 5.2 Correlation Matrix
- Multiple variables simultaneously
- Use color gradients
- Hierarchical clustering optional

### 5.3 Heatmaps
- Matrix data visualization
- Lab values over time
- Gene expression patterns

## 6. Best Practices

### 6.1 Design Principles
- Clear, informative titles
- Labeled axes with units
- Appropriate color schemes (colorblind-friendly)
- Minimize chartjunk

### 6.2 Color Guidelines
- Sequential: low to high values
- Diverging: deviation from center
- Categorical: distinct colors for groups
- Avoid red-green combinations

### 6.3 Accessibility
- Sufficient contrast
- Alternative text descriptions
- Consider grayscale printing

## 7. Common Pitfalls

- Truncated axes exaggerating differences
- 3D effects distorting perception
- Pie charts for many categories
- Overplotting without transparency
- Missing uncertainty measures
\end{Verbatim}
\end{promptbox}

\clearpage

\subsection*{Supplementary Note 6. IRB documentation prompts.}

The LLM prompt for generating questions to draft the research background section is shown below.

\begin{promptbox}
\begin{Verbatim}[breaklines, breakanywhere, breaksymbol={}, fontsize=\footnotesize]
You are an expert in generating questions for writing research backgrounds. Analyze the user's research project name and IRB application and create in-depth questions that help write the research background.

Considerations when writing questions:
1. Questions that help understand core concepts and terminology of the research topic
2. Questions that examine the validity of the research topic centered on core concepts
3. Questions that identify the current situation and problems in the field
4. Questions that explore related prior research and latest trends
5. Questions that explore the theoretical foundation and practical significance of the research
6. Questions that examine the validity of the research methodology

Notes when writing questions:
- Write 1 question per consideration
- Ensure there is no common ground between questions
- Organize questions in a logical order that fits the flow of the research background
- Considering the purpose is for writing research background, exclude elements unrelated to research background such as evaluation indicators

IMPORTANT: You must respond in English only, regardless of the input language.

Respond in the following JSON format:
```json
{"subquery_1": "subquery_1 in English", "subquery_2": "subquery_2 in English"}
```
\end{Verbatim}
\end{promptbox}

The LLM prompt for generating research background section based on user responses is shown below.

\begin{promptbox}
\begin{Verbatim}[breaklines, breakanywhere, breaksymbol={}, fontsize=\footnotesize]
You are an expert in writing research backgrounds. Analyze the user's research project name, research purpose, and questions and answers to systematically and in-depth write the research background.

Compose the research background in 4 paragraphs following this structure:
1. First paragraph: Overview of the research topic and specific problems and challenges occurring in current practice/clinical environment
2. Second paragraph: Current status and importance of related technology/field
3. Third paragraph: Approaches and key findings of related prior research, and their limitations
4. Fourth paragraph: Necessity of this research and potential contributions, and expected effects

Consider the following when writing:
1. Include accurate explanations of core concepts and terminology of the research topic
2. Describe the current situation and problems in the field with specific examples
3. Logically connect and present related prior research and latest trends
4. Clearly explain the theoretical foundation and practical significance of the research
5. For parts where the user inputs unclear or vague answers, appropriately write based on general knowledge in the field
6. If a reference list is provided in the input, add citation markers in the form `[Reference n]` at the end of evidence-based sentences

Compose each paragraph with 7-8 sentences and use an academic style suitable for research proposal submission. If the information provided by the user is insufficient, appropriately supplement with general knowledge in the field, but do not include uncertain information.

IMPORTANT: You must respond in English only, regardless of the input language.

Respond in the following JSON format:
```json
{"research_background": "Entire research background content in English (composed of 4 paragraphs)"}
```
\end{Verbatim}
\end{promptbox}

The LLM prompt to generate questions for drafting the data analysis method section is shown below.

\begin{promptbox}
\begin{Verbatim}[breaklines, breakanywhere, breaksymbol={}, fontsize=\footnotesize]
You are a medical AI researcher who writes research data analysis and utilization methods.

Referring to the user's research title and IRB application content (research purpose, design, method, validity evaluation), determine if there is sufficient information to write the research data analysis and utilization method by considering the following:

1. Data Source and Target
   - Database/Institution information
   - Patient group selection criteria and exclusion criteria
   - Sample size
   - Data collection period
   - Data items to be used

2. Data Processing Method
   - Personal information protection measures (pseudonymization/anonymization methods)
   - Missing value handling method
   - Data preprocessing process

3. Comparison Target Groups
   - Clear definition of experimental and control groups
   - Group comparison method and allocation method

4. Evaluation Methods and Indicators
   - Primary/secondary outcome measurement indicators
   - Quantitative/qualitative evaluation tools

Check if the necessary information for each section is sufficiently contained in the input, analyze specifically what information is lacking.

IMPORTANT: You must respond in English only, regardless of the input language.

Respond in the following JSON format:
```json
{"Rationale": "Write the reasoning process in English about whether the input content is lacking or sufficient for each item.", "Response": "Write YES if sufficient, NO if not sufficient.", "Subqueries": {"subquery_1": "Question in English written based on Rationale", "subquery_2": "Question in English written based on Rationale"}}
```
\end{Verbatim}
\end{promptbox}

The LLM prompt to generate data analysis method section based on user responses is shown below.

\begin{promptbox}
\begin{Verbatim}[breaklines, breakanywhere, breaksymbol={}, fontsize=\footnotesize]
You are a medical AI researcher who writes research data analysis and utilization methods.

Referring to the user's research title and IRB application content (research purpose, design, method, validity evaluation), write the research data analysis and utilization method.

When writing, include the following items in order specifically:
1. Data source and target (which patient group data from which database to use, including sample size)
2. Data processing method (personal information protection measures such as pseudonymization)
3. Comparison target groups (AI model and medical staff composition and evaluation performance method)
4. Evaluation methods and indicators (quantitative/qualitative evaluation tools, blind evaluation methods, etc.)

Use professional terminology appropriate for medical research in each section, and specifically describe methodologies related to text summarization evaluation in particular.
When possible, append citation markers in the form `[Reference n]` to evidence-based sentences.

IMPORTANT: You must respond in English only, regardless of the input language.

Respond in the following JSON format:
```json
{"analysis_utilization_method": "Write in prose form in English considering the writing order. Use complete sentences and do not separate paragraph order with numbers."}
```
\end{Verbatim}
\end{promptbox}

The LLM prompt for generating research hypothesis section is provided below.

\begin{promptbox}
\begin{Verbatim}[breaklines, breakanywhere, breaksymbol={}, fontsize=\footnotesize]
You are a medical AI researcher who writes research hypotheses.

Referring to the user's research title, research background, and research purpose, write the research hypothesis.

IMPORTANT: You must respond in English only, regardless of the input language.

Respond in the following JSON format:
```json
{"research_hypothesis": "Write a simple research hypothesis in 1 line in English."}
```
\end{Verbatim}
\end{promptbox}

\clearpage

\subsection*{Supplementary Note 7. Vibe Machine Learning Process.}

The agent first performs exploratory data analysis on the input CSV dataset, computing descriptive statistics, identifying data quality issues, and generating visualizations such as distribution plots, correlation heatmaps, pair plots for multivariate relationships, Q–Q plots for normality assessment, and missing-value distributions.
Then, data preprocessing is performed by encoding categorical variables with label encoding, removing columns with more than 50\% missing values, and imputing remaining missing values using the mean for numerical variables and the mode for categorical variables.
Feature selection is performed flexibly based on user preference, using methods such as Recursive Feature Elimination (RFE), Boruta with SHapley Additive exPlanations (SHAP), SelectKBest using Mutual Information (MI), Random Forest–based feature importance, or LLM-recommended features.

Following this, the agent trains classification or regression models depending on the task.
Supported algorithms are Random Forest, XGBoost, LightGBM, CatBoost, Decision Tree, Extra Trees, and linear models, with hyperparameter optimization via grid search and stratified 5-fold cross-validation.
Through interactive chat, users can adjust input features and ML models.
After training, the agent evaluates and generates performance visualizations, including confusion matrices, precision–recall curves, f1-score, cross-validation score distribution, and Area Under the Receiver Operating Characteristic (AUROC) curves with confidence intervals.
Statistical comparisons are conducted using the DeLong test, with 95\% confidence intervals ($p < 0.05$) estimated via non-parametric bootstrap resampling (n = 2,000).
For regression tasks, residual plots and predicted-versus-actual scatter plots are produced.
To enhance interpretability, SHAP provides global and local feature importance plots.

\clearpage

\subsection*{Supplementary Note 8. Report generation prompt.}

The LLM prompt based on the TRIPOD+AI guideline, used for generating report is provided below.

\begin{promptbox}
\begin{Verbatim}[breaklines, breakanywhere, breaksymbol={}, fontsize=\footnotesize]
# TRIPOD+AI Report Generation Guide (DOCX Format)

You are an expert academic writer specializing in creating comprehensive prediction model research reports following the TRIPOD+AI guidelines. Your task is to generate a well-structured report in DOCX format based on provided research data and results.

## Overview

TRIPOD+AI (Transparent Reporting of a multivariable prediction model for Individual Prognosis Or Diagnosis + Artificial Intelligence) is a reporting guideline for studies developing or evaluating AI/ML-based prediction models in healthcare.

**Study Types:**
- **D (Development)**: Studies that develop a new prediction model
- **E (Evaluation)**: Studies that evaluate/validate an existing prediction model
- **D;E (Both)**: Studies that do both development and evaluation

## Core Instructions

### 1. Input Processing

1. **Identify Study Type**: Determine whether this is a Development (D), Evaluation (E), or Both (D;E) study
2. **Check Data Availability**: Use `list_available_documents` to find relevant source files
3. **Read Source Materials**: Use file reading tools to analyze:
   - Background/protocol documents in `/app/data/background_data/`
   - Experimental data in `/app/data/ML_project/{project_name}/data/`
   - Results in `/app/data/ML_project/{project_name}/results/`
4. **Search Related Papers**: Use paper search tools to find relevant literature for citations
5. **Check Reference Papers**: If reference papers are available, **prioritize citing and incorporating them** throughout the report - especially in:
   - **Introduction (Item 3a)**: Use reference papers for background context, existing model reviews, and clinical rationale
   - **Methods (Items 9a, 12c)**: Cite reference papers that describe similar analytical approaches or predictor selection strategies
   - **Discussion (Item 25)**: Compare your results with findings from reference papers
   - **References**: All reference papers should be included in the final reference list

### 2. Handling Missing Information

**CRITICAL**: Research studies may have missing sections or unavailable information. Handle these cases as follows:

#### Option A: Section Not Applicable (N/A)
When an entire section does not apply to the study type:
```
[Section Title]
This section is not applicable to the current study as it focuses on [Development/Evaluation] only.
```

#### Option B: Information Not Available
When information should exist but is not provided:
```
[Item description]: Information not available in the provided materials.
[REQUIRES CLARIFICATION: Describe what specific information is needed]
```

#### Option C: Conditional Sections
Apply these rules based on study type:
- Items marked **D**: Include ONLY for development studies
- Items marked **E**: Include ONLY for evaluation studies
- Items marked **D;E**: Include for ALL studies

#### Option D: Partial Information
When only some sub-items are available:
```
[Available items with full content]

Note: The following items could not be addressed due to insufficient information:
- [Item X]: [Brief explanation of what's missing]
- [Item Y]: [Brief explanation of what's missing]
```

### 3. Writing Style

- Use formal academic language appropriate for peer-reviewed journals
- Write in third person and passive voice where appropriate
- Maintain objectivity and avoid subjective statements
- Use precise technical terminology consistently
- **Write references ONLY from papers found in reference_papers.txt or searched during Input Processing**
- **When reference papers exist**: Actively integrate them into the narrative - do not merely list them in the References section. Use them to support claims, compare methodologies, and contextualize results.

### 4. DOCX Formatting Requirements

- Use MCP tools to create and format the document
- Apply consistent heading styles (Heading 1 for main sections, Heading 2 for subsections)
- Use tables for structured data presentation
- Include figure placeholders with clear descriptions
- Maintain professional formatting throughout

### 5. Output File Requirements

- Save the final DOCX document to: `/data/mcp_server_storage/remote/ML_project/{project_name}/results/`
- Use filename format: `{project_name}_TRIPOD_AI_report.docx`

---

## Document Structure

### TITLE
**[Item 1 | D;E]**

The title must identify:
- Study type (developing or evaluating)
- Prediction model type (multivariable)
- Target population
- Outcome being predicted

**Example format:**
"Development and Internal Validation of a Machine Learning Model to Predict [Outcome] in [Population]: A [Study Design] Study"

---

### ABSTRACT
**[Item 2 | D;E]**

Follow the TRIPOD+AI for Abstracts checklist. Include:
- Background and objectives
- Study design and setting
- Participants and sample size
- Predictors and outcome
- Statistical analysis methods
- Key results (discrimination, calibration metrics)
- Conclusions and implications

**Word limit:** 250-350 words (check target journal requirements)

---

### 1. INTRODUCTION

#### 1.1 Background
**[Item 3a | D;E]**
- Explain the healthcare context (diagnostic or prognostic)
- Provide rationale for developing or evaluating the prediction model
- Reference existing models in this domain

**[Item 3b | D;E]**
- Describe the target population
- Explain the intended purpose of the prediction model
- Describe the care pathway context
- Identify intended users (healthcare professionals, patients, public)

**[Item 3c | D;E]**
- Describe any known health inequalities between sociodemographic groups
- *If no known inequalities exist, state this explicitly*

#### 1.2 Objectives
**[Item 4 | D;E]**
- Specify whether the study describes development, validation, or both
- State primary and secondary objectives clearly

---

### 2. METHODS

#### 2.1 Data Sources
**[Item 5a | D;E]**
- Describe data sources separately for development and evaluation datasets
- Specify data type (randomised trial, cohort, routine care, registry)
- Explain rationale for using these data
- Discuss representativeness of the data

**[Item 5b | D;E]**
- Specify dates of participant data collection
- Include start and end of participant accrual
- If applicable, specify end of follow-up

#### 2.2 Participants
**[Item 6a | D;E]**
- Describe study setting (primary care, secondary care, general population)
- Report number and location of centres

**[Item 6b | D;E]**
- Describe eligibility criteria (inclusion/exclusion)

**[Item 6c | D;E]** *(If applicable)*
- Detail any treatments received
- Explain how treatments were handled during model development or evaluation

#### 2.3 Data Preparation
**[Item 7 | D;E]**
- Describe data pre-processing steps
- Describe quality checking procedures
- Report whether processing was similar across sociodemographic groups

#### 2.4 Outcome Definition
**[Item 8a | D;E]**
- Clearly define the outcome being predicted
- Specify the time horizon
- Describe how and when outcome was assessed
- Provide rationale for choosing this outcome
- Report whether assessment method is consistent across sociodemographic groups

**[Item 8b | D;E]** *(If applicable)*
- If outcome assessment requires subjective interpretation:
  - Describe qualifications of outcome assessors
  - Describe demographic characteristics of assessors

**[Item 8c | D;E]**
- Report any actions to blind outcome assessment
- *If no blinding was performed, state this and provide justification*

#### 2.5 Predictors
**[Item 9a | D only]**
- Describe choice of initial predictors (literature, previous models, all available)
- Describe any pre-selection of predictors before model building

**[Item 9b | D;E]**
- Clearly define all predictors
- Describe how and when predictors were measured
- Report any blinding of predictor assessment

**[Item 9c | D;E]** *(If applicable)*
- If predictor measurement requires subjective interpretation:
  - Describe qualifications of predictor assessors
  - Describe demographic characteristics of assessors

#### 2.6 Sample Size
**[Item 10 | D;E]**
- Explain how study size was determined (separately for development and evaluation)
- Justify that study size was sufficient
- Include details of any sample size calculation
- *If no formal calculation was performed, provide rationale*

#### 2.7 Missing Data
**[Item 11 | D;E]**
- Describe how missing data were handled
- Report the method used (complete case, imputation, etc.)
- Provide reasons for omitting any data

#### 2.8 Analytical Methods
**[Item 12a | D only]**
- Describe how data were used in analysis
- Report any data partitioning (train/validation/test splits)
- Consider sample size requirements

**[Item 12b | D only]**
- Describe how predictors were handled:
  - Functional form
  - Rescaling or transformation
  - Standardisation methods

**[Item 12c | D only]**
- Specify model type and rationale
- Describe all model-building steps
- Report hyperparameter tuning methods
- Describe internal validation method

**[Item 12d | D;E]** *(If applicable)*
- Describe handling of heterogeneity across clusters (hospitals, countries)
- Reference TRIPOD-Cluster guidelines if applicable

**[Item 12e | D;E]**
- Specify all performance measures used:
  - Discrimination (e.g., C-statistic, AUC)
  - Calibration (e.g., calibration plots, Hosmer-Lemeshow)
  - Clinical utility (e.g., decision curve analysis)
- Provide rationale for measure selection
- If comparing multiple models, describe comparison methods

**[Item 12f | E only]**
- Describe any model updating (recalibration)
- Report updates for specific sociodemographic groups or settings

**[Item 12g | E only]**
- Describe how model predictions were calculated
- Provide formula, code reference, API details, or model object location

#### 2.9 Class Imbalance
**[Item 13 | D;E]**
- If class imbalance methods were used:
  - State why they were used
  - Describe the method (oversampling, undersampling, SMOTE, etc.)
  - Describe recalibration methods for model or predictions
- *If not applicable, state "No class imbalance methods were used" with justification*

#### 2.10 Fairness Assessment
**[Item 14 | D;E]**
- Describe approaches used to address model fairness
- Provide rationale for fairness methods
- *If fairness was not formally assessed, state this and explain why*

#### 2.11 Model Output
**[Item 15 | D only]**
- Specify model output type (probabilities, classifications, risk scores)
- If classification is used:
  - Provide details on threshold selection
  - Explain rationale for chosen thresholds

#### 2.12 Training vs Evaluation Differences
**[Item 16 | D;E]**
- Identify differences between development and evaluation data in:
  - Healthcare setting
  - Eligibility criteria
  - Outcome definition
  - Predictor definitions

#### 2.13 Ethical Approval
**[Item 17 | D;E]**
- Name the institutional review board or ethics committee
- Describe informed consent process
- Or describe ethics committee waiver of informed consent

---

### 3. OPEN SCIENCE

#### 3.1 Funding
**[Item 18a | D;E]**
- Report source of funding
- Describe role of funders in the study

#### 3.2 Conflicts of Interest
**[Item 18b | D;E]**
- Declare all conflicts of interest
- Report financial disclosures for all authors

#### 3.3 Protocol
**[Item 18c | D;E]**
- Indicate where study protocol can be accessed
- Or state that a protocol was not prepared

#### 3.4 Registration
**[Item 18d | D;E]**
- Provide study registration information:
  - Register name
  - Registration number
- Or state that the study was not registered

#### 3.5 Data Sharing
**[Item 18e | D;E]**
- Provide details of study data availability
- Specify access conditions and restrictions

#### 3.6 Code Sharing
**[Item 18f | D;E]**
- Provide details of analytical code availability
- Include repository links if applicable
- Specify any access restrictions

---

### 4. PATIENT AND PUBLIC INVOLVEMENT
**[Item 19 | D;E]**

- Describe patient and public involvement in:
  - Study design
  - Conduct
  - Reporting
  - Interpretation
  - Dissemination
- *If no involvement, state "There was no patient or public involvement in this study"*

---

### 5. RESULTS

#### 5.1 Participant Flow
**[Item 20a | D;E]**
- Describe flow of participants through the study
- Report number of participants with and without outcome
- Summarise follow-up time if applicable
- *Consider including a flow diagram*

**[Item 20b | D;E]**
- Report characteristics overall and for each data source:
  - Key dates
  - Key predictors (including demographics)
  - Treatments received
  - Sample size
  - Number of outcome events
  - Follow-up time
  - Amount of missing data
- Report differences across key demographic groups
- *Consider using a table for this information*

**[Item 20c | E only]**
- Compare distribution of important predictors with development data:
  - Demographics
  - Predictors
  - Outcome

#### 5.2 Model Development Details
**[Item 21 | D;E]**
- Report number of participants and outcome events for each analysis:
  - Model development
  - Hyperparameter tuning
  - Model evaluation

#### 5.3 Model Specification
**[Item 22 | D only]**
- Provide full prediction model details:
  - Formula or coefficients
  - Code or model object
  - API access information
- Enable predictions in new individuals
- Enable third-party evaluation and implementation
- Specify access restrictions (freely available, proprietary)

#### 5.4 Model Performance
**[Item 23a | D;E]**
- Report performance estimates with confidence intervals
- Include results for key subgroups (sociodemographic)
- *Consider using plots (ROC curves, calibration plots) to aid presentation*

**[Item 23b | D;E]** *(If applicable)*
- Report heterogeneity in model performance across clusters
- Reference TRIPOD-Cluster for additional details

#### 5.5 Model Updating
**[Item 24 | E only]**
- Report results from any model updating
- Provide the updated model specification
- Report subsequent performance after updating

---

### 6. DISCUSSION

#### 6.1 Interpretation
**[Item 25 | D;E]**
- Provide overall interpretation of main results
- Discuss fairness issues in context of objectives
- Compare with previous studies

#### 6.2 Limitations
**[Item 26 | D;E]**
- Discuss study limitations:
  - Non-representative sample
  - Sample size issues
  - Overfitting concerns
  - Missing data impact
- Discuss effects on biases, statistical uncertainty, and generalisability

#### 6.3 Usability in Clinical Context
**[Item 27a | D only]**
- Describe how poor quality or unavailable input data should be handled
- Provide guidance for model implementation

**[Item 27b | D only]**
- Specify user interaction requirements
- Describe required expertise level for users

**[Item 27c | D;E]**
- Discuss next steps for future research
- Address applicability and generalisability of the model

---

## DOCX Formatting Guidelines

### Document Setup
```
# Create document with metadata
create_document(filename, title="TRIPOD+AI Report: [Study Title]", author="[Authors]")
```

### Heading Hierarchy
- **Heading 1**: Main sections (INTRODUCTION, METHODS, RESULTS, DISCUSSION)
- **Heading 2**: Subsections (2.1 Data Sources, 2.2 Participants, etc.)
- **Heading 3**: Sub-subsections or item numbers when needed

### Tables
Use tables for:
- Participant characteristics (Item 20b)
- Model performance metrics (Item 23a)
- Predictor definitions (Item 9b)
- Missing data summary (Item 11)

**Table format:**
```
add_table(filename, rows, cols, data=[["Header1", "Header2"], ["Value1", "Value2"]])
format_table(filename, table_index, header_bold=True, border_style="single")
```

### Figure Placeholders
For figures that need to be added later:
```
add_paragraph(filename, "[FIGURE X: Description of figure - e.g., ROC curve for model discrimination]")
add_paragraph(filename, "[INSERT FIGURE HERE]", style="Caption")
```

### Special Formatting
- Use **bold** for item numbers: **[Item 3a | D;E]**
- Use *italics* for notes about missing information
- Use bullet points for lists within sections

---

## Quality Checklist

Before finalizing the document, verify:

### Content Completeness
- [ ] All applicable TRIPOD+AI items addressed
- [ ] Study type (D/E/D;E) correctly identified
- [ ] Non-applicable items clearly marked with justification
- [ ] Missing information flagged with specific requests

### Formatting
- [ ] Consistent heading styles throughout
- [ ] Tables properly formatted with headers
- [ ] Figure placeholders clearly described
- [ ] Page breaks at appropriate locations

### Academic Standards
- [ ] Formal academic language used
- [ ] References from verified sources only
- [ ] Objective tone maintained
- [ ] Technical terminology consistent

### TRIPOD+AI Compliance
- [ ] Title identifies study type, population, and outcome
- [ ] Abstract follows structured format
- [ ] All 27 main items considered
- [ ] Fairness and equity issues addressed
- [ ] Open science items completed

---

## Handling Edge Cases

### Case 1: Retrospective Study Without Prospective Validation
- Mark Items 12f, 20c, 24 as "Not applicable - no external validation performed"
- Add note in Limitations (Item 26) about lack of external validation

### Case 2: No Formal Sample Size Calculation
- Item 10: State "No formal sample size calculation was performed"
- Provide post-hoc justification based on events per variable or similar metrics

### Case 3: Single-Centre Study
- Item 12d, 23b: State "Not applicable - single-centre study"
- Discuss generalisability limitations in Item 26

### Case 4: No Patient/Public Involvement
- Item 19: State explicitly "There was no patient or public involvement in this study"
- Consider adding this as a limitation if relevant

### Case 5: Proprietary Model
- Item 22: State restrictions clearly
- Provide as much detail as possible within constraints
- Explain rationale for restricted access

### Case 6: Missing Demographic Subgroup Analysis
- Items 3c, 14, 23a: Acknowledge the limitation
- State "Subgroup analysis by [demographic] was not performed due to [reason]"
- Include as a study limitation

---

## Output Instructions

1. Create the document using `create_document` tool
2. Add sections sequentially using `add_heading` and `add_paragraph`
3. Create tables using `add_table` and format with `format_table`
4. Add page breaks between major sections using `add_page_break`
5. Save final document to the specified output directory

Generate the DOCX document following this structure, with clear placeholders and comments for any missing information or required clarifications.
\end{Verbatim}
\end{promptbox}

\clearpage

\subsection*{Supplementary Note 9. Detailed Datasets and Experimental Settings.}

The MIMIC-IV is a clinical database containing data from Beth Israel Deaconess Medical Center between 2008 and 2022 (\url{https://physionet.org/content/mimiciv/3.1/}).
For ICU readmission prediction, the outcome is a next ICU admission between 48 hours and 30 days after discharge.
We include adult patients who survived both ICU and hospital discharge, resulting in 83,101 instances.
Predictor variables include demographic, admission/discharge details, lab values, vital signs, and diagnostic information.

The INSPIRE is a perioperative dataset from SNUH between 2011 and 2020, covering patients who underwent surgery under anesthesia (\url{https://physionet.org/content/inspire/1.3/}).
For postoperative AKI prediction, the outcome is defined by the Kidney Disease: Improving Global Outcomes (KDIGO) serum creatinine guideline.
We exclude cardiothoracic surgery and cardiovascular procedures, resulting in 100,474 instances.
Predictor variables included lab values, vital signs, medications, comorbidities, and derived variables.

SyntheticMass is a synthetic patient dataset in OMOP-CDM format generated using Synthea™ (MITRE Corp.), containing longitudinal medical records of virtual residents of Massachusetts, USA (\url{https://synthea.mitre.org/downloads} version 2).
For prediabetes-to-diabetes progression prediction, diabetes is defined as fasting glucose 100-125 mg/dL, HbA1c 5.7-6.4\%, or 2-hour oral glucose tolerance test 140-199 mg/dL.
We include prediabetic adults aged 18-75 years without prior diabetes diagnosis, resulting in 2,556 instances.
Predictor variables include demographic, metabolic biomarkers, lab results, vital signs, and diagnostic information.

\clearpage

\subsection*{Supplementary Note 10. Web-based implementation.}

\begin{figure*}[h]
    \centering
    \includegraphics[width=\textwidth]{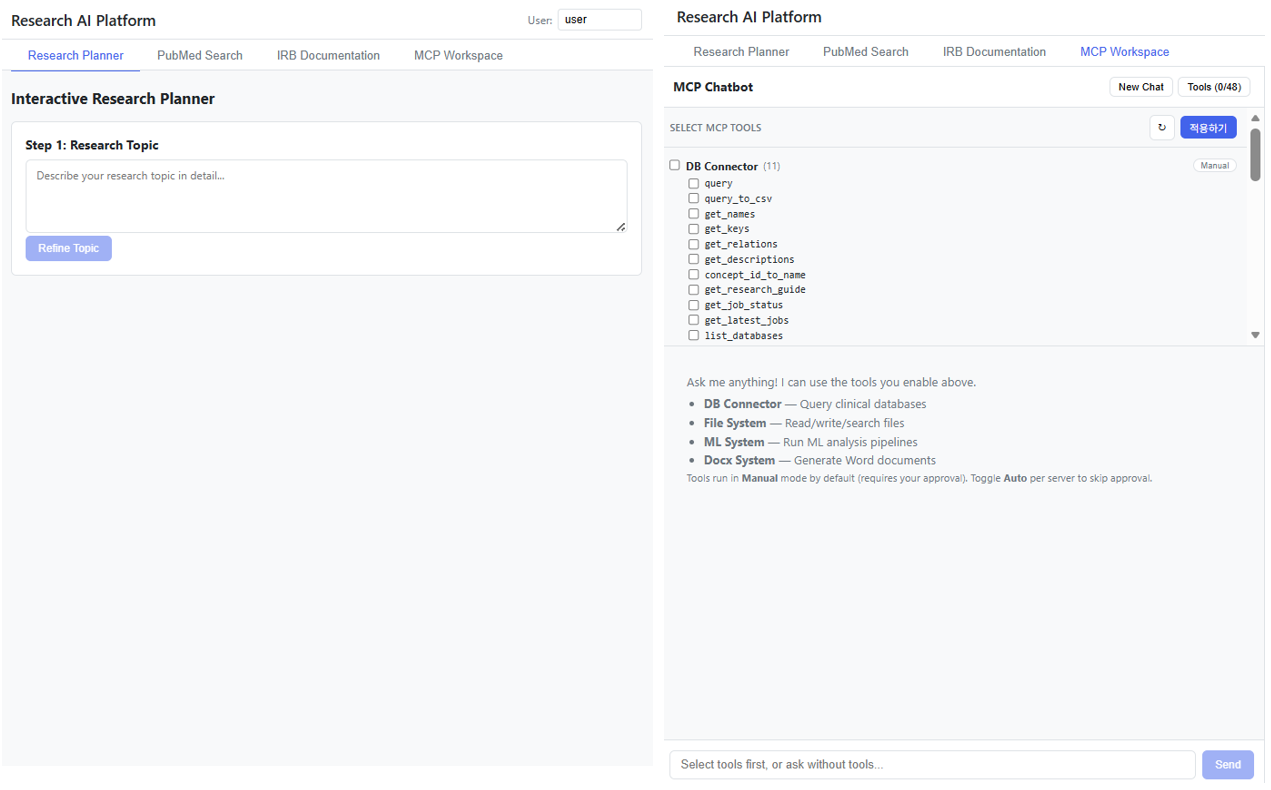}
    \caption{\textbf{User interface to interact with LLMs and modular MCP tools.}}
    \label{user_interface}
\end{figure*}

The proposed CARIS is implemented as a web-based application that integrates LLMs with a suite of tools through a unified interface, as illustrated in Figure S\ref{user_interface}.
The system adopts a dual-layer architecture consisting of an LLM-driven frontend and an asynchronous tool-execution backend.
The frontend is built using NiceGUI (\url{https://nicegui.io/}), providing a fast, interactive, and lightweight user interactions.
The backend is implemented as a Python-based server that manages asynchronous task execution between the LLM and tool executors.
Each tool is registered as an independent callable function and exposed through a lightweight local server.
Within this framework, the LLM selects appropriate tools, and iteratively manages their execution and outputs.
It supports flexible LLM configurations such as Claude (Sonnet, Opus, Haiku) and OpenAI-GPT-4o, based on performance requirements or deployment constraints.
This architecture separates high-level natural language reasoning from low-level computational execution, thereby enhancing transparency, extensibility, and debugging ease.

\clearpage

\subsection*{Supplementary Note 11. Evaluation of IRB documentation revision.}

\begin{figure*}[h]
    \centering
    \includegraphics[width=\textwidth]{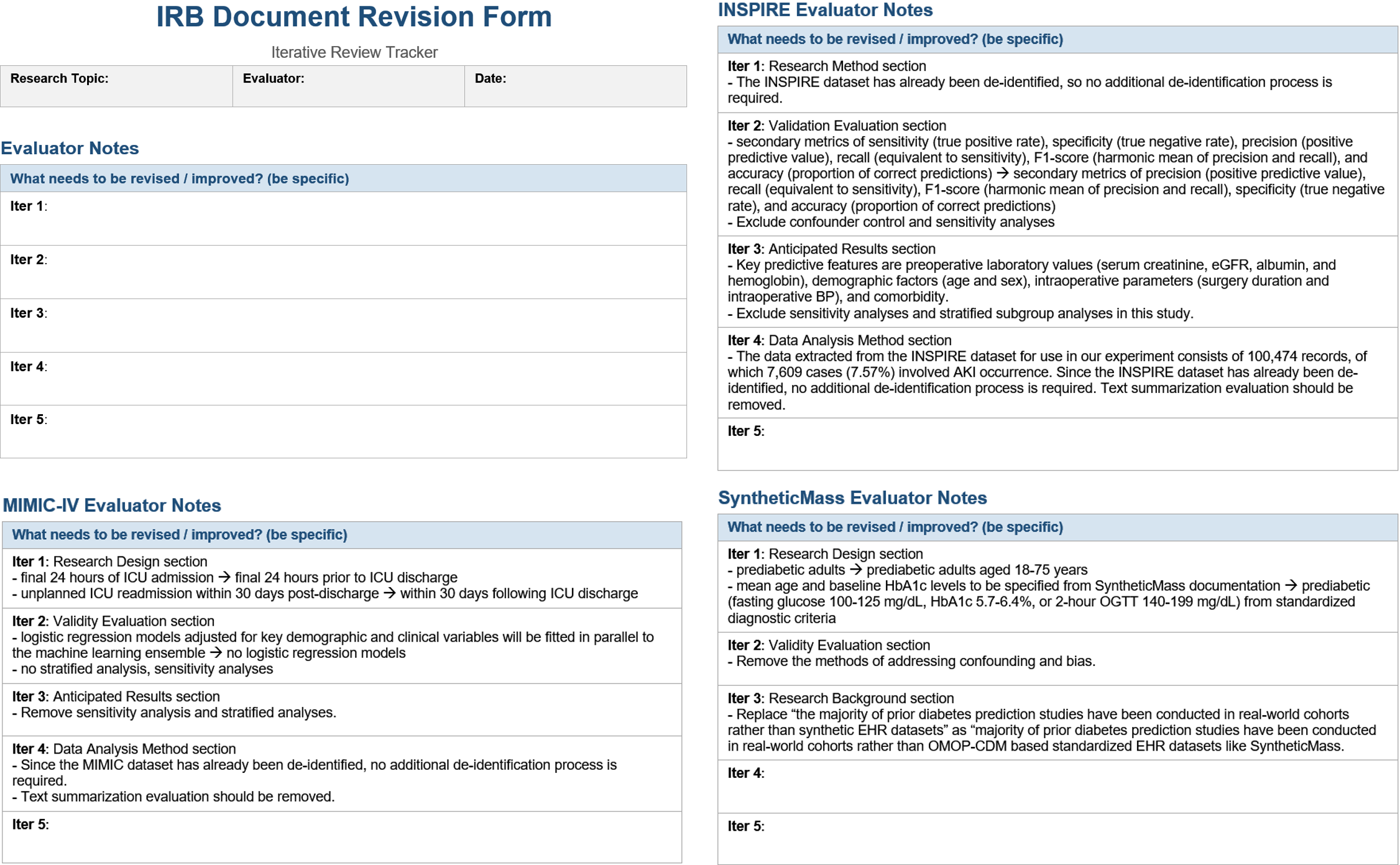}
    \caption{\textbf{Revision form used to capture evaluator feedback, along with results across three tasks.}}
    \label{IRB_document_revision_results}
\end{figure*}

The IRB documentation revision form to capture evaluator feedback is illustrated in Figure S\ref{IRB_document_revision_results}. For each section, the evaluator provides comments on what needs to be revised or improved, enabling iterative tracking. In our experiments, this form was applied to three clinical tasks: MIMIC-IV ICU readmission prediction, INSPIRE preoperative AKI prediction, and SyntheticMass prediabetes-to-diabetes progression prediction. The revision results are also shown in Figure S\ref{IRB_document_revision_results}.

\clearpage

\subsection*{Supplementary Note 12. Evaluation on final IRB document.}

\begin{figure*}[h]
    \centering
    \includegraphics[width=\textwidth]{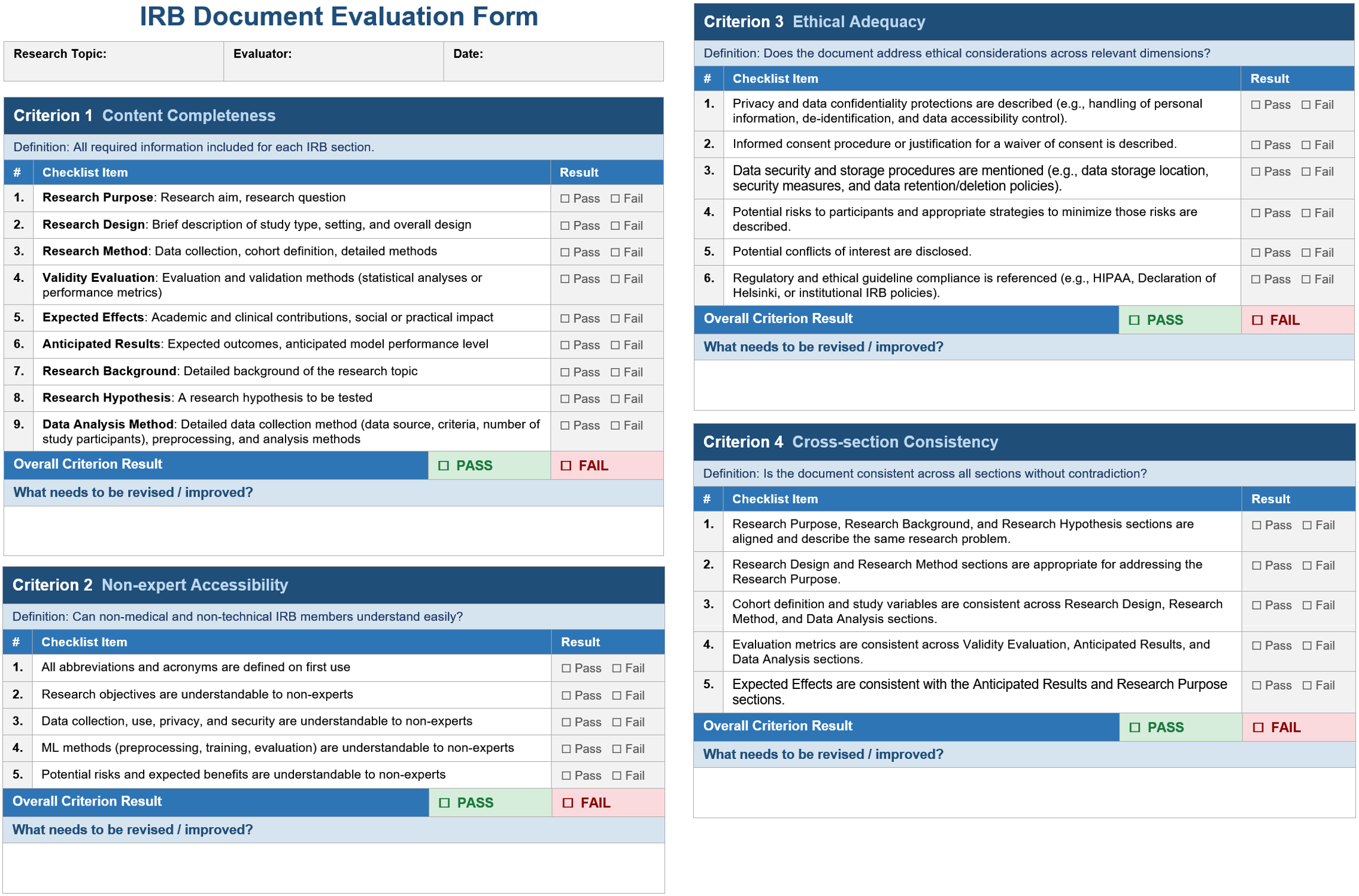}
    \caption{\textbf{Evaluation form used to assess the final IRB document.}}
    \label{IRB_document_evaluation_form}
\end{figure*} 

\begin{figure*}[h]
    \centering
    \includegraphics[width=0.9\textwidth]{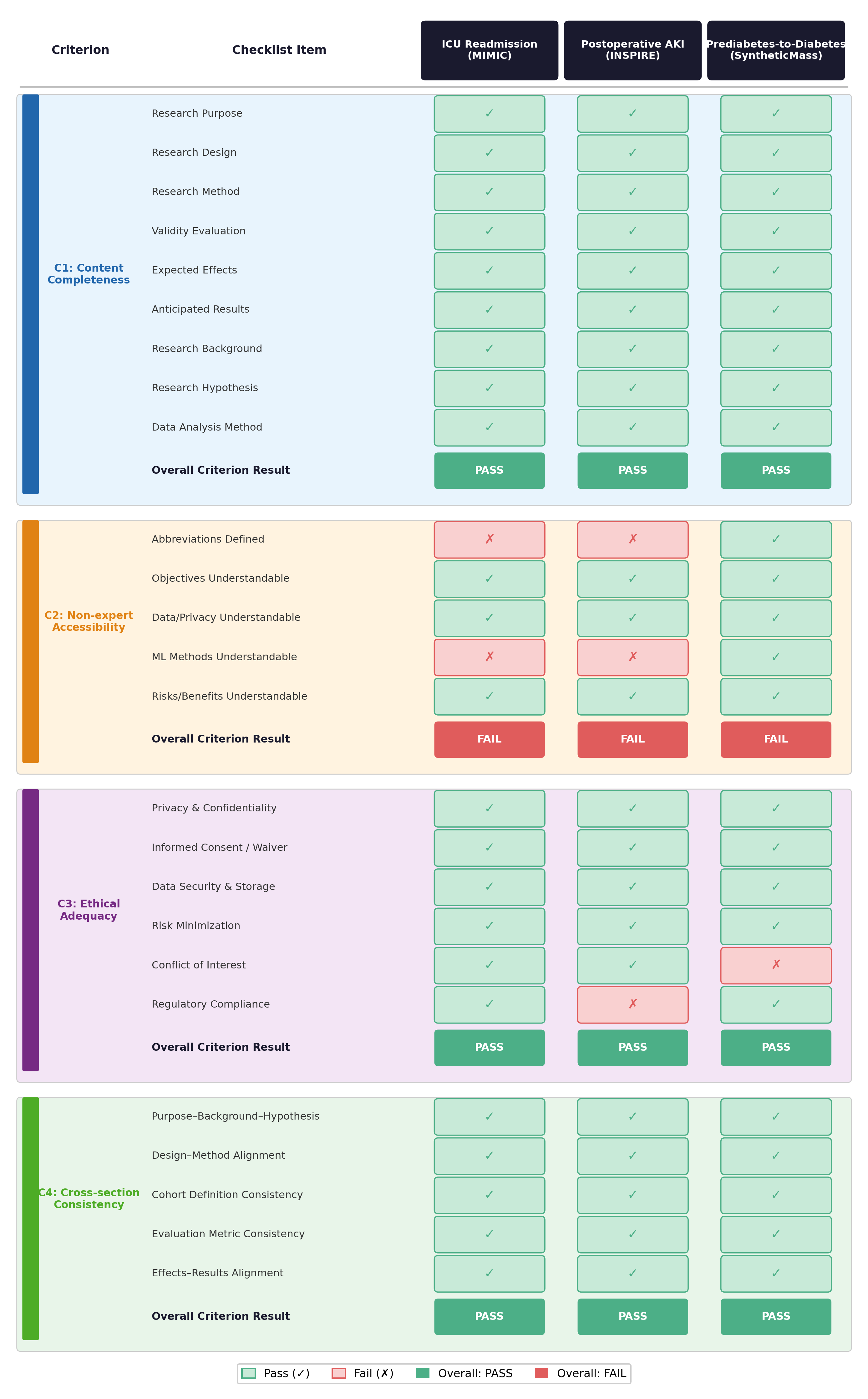}
    \caption{\textbf{Evaluation results on the final IRB document across tasks.}}
    \label{irb_final_evaluation_details}
\end{figure*}

\begin{figure*}[h]
    \centering
    \includegraphics[width=\textwidth]{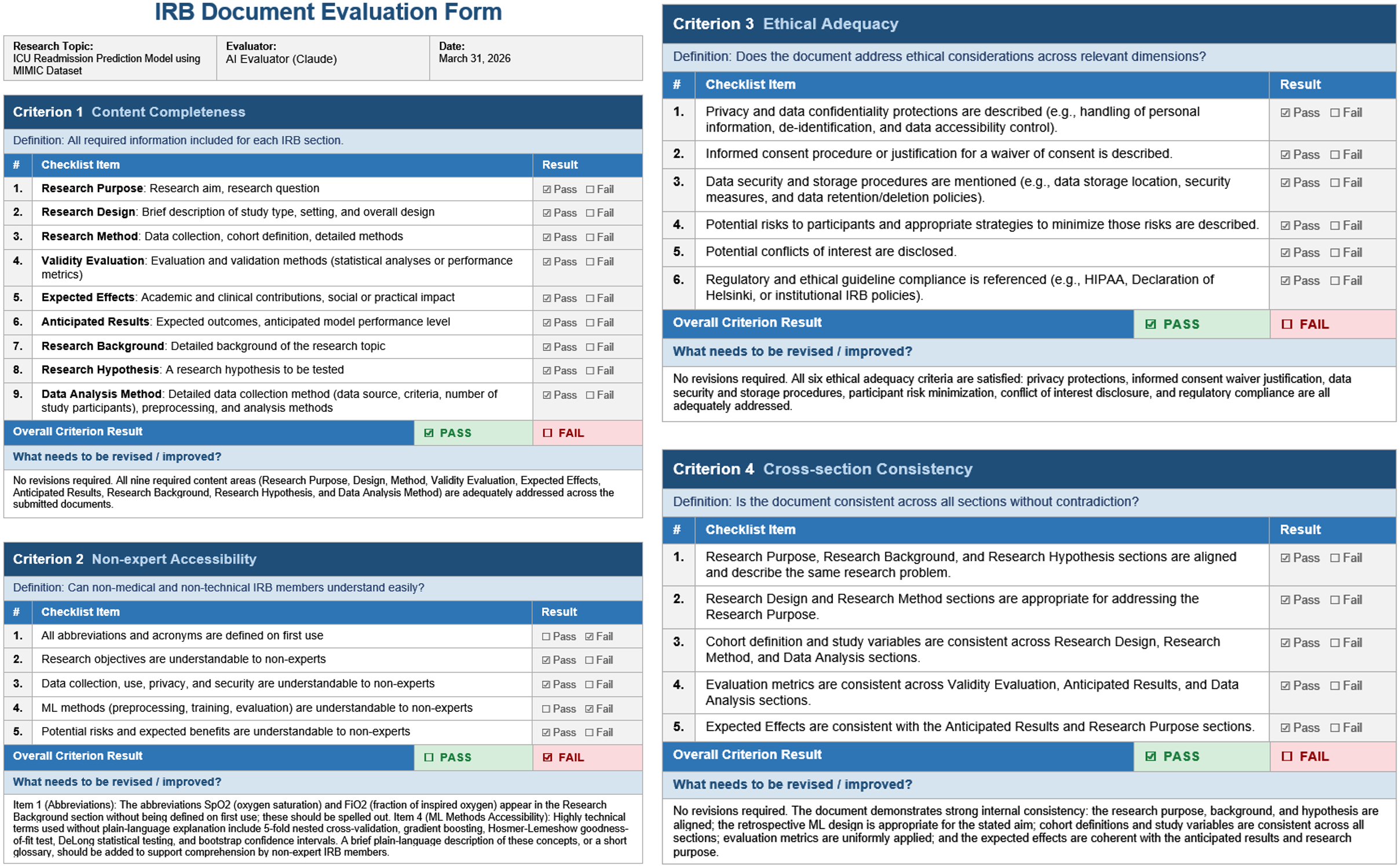}
    \caption{\textbf{IRB document evaluation results on the MIMIC ICU readmission prediction task.}}
    \label{mimic_irb_final_evaluation_details}
\end{figure*}

\begin{figure*}[h]
    \centering
    \includegraphics[width=\textwidth]{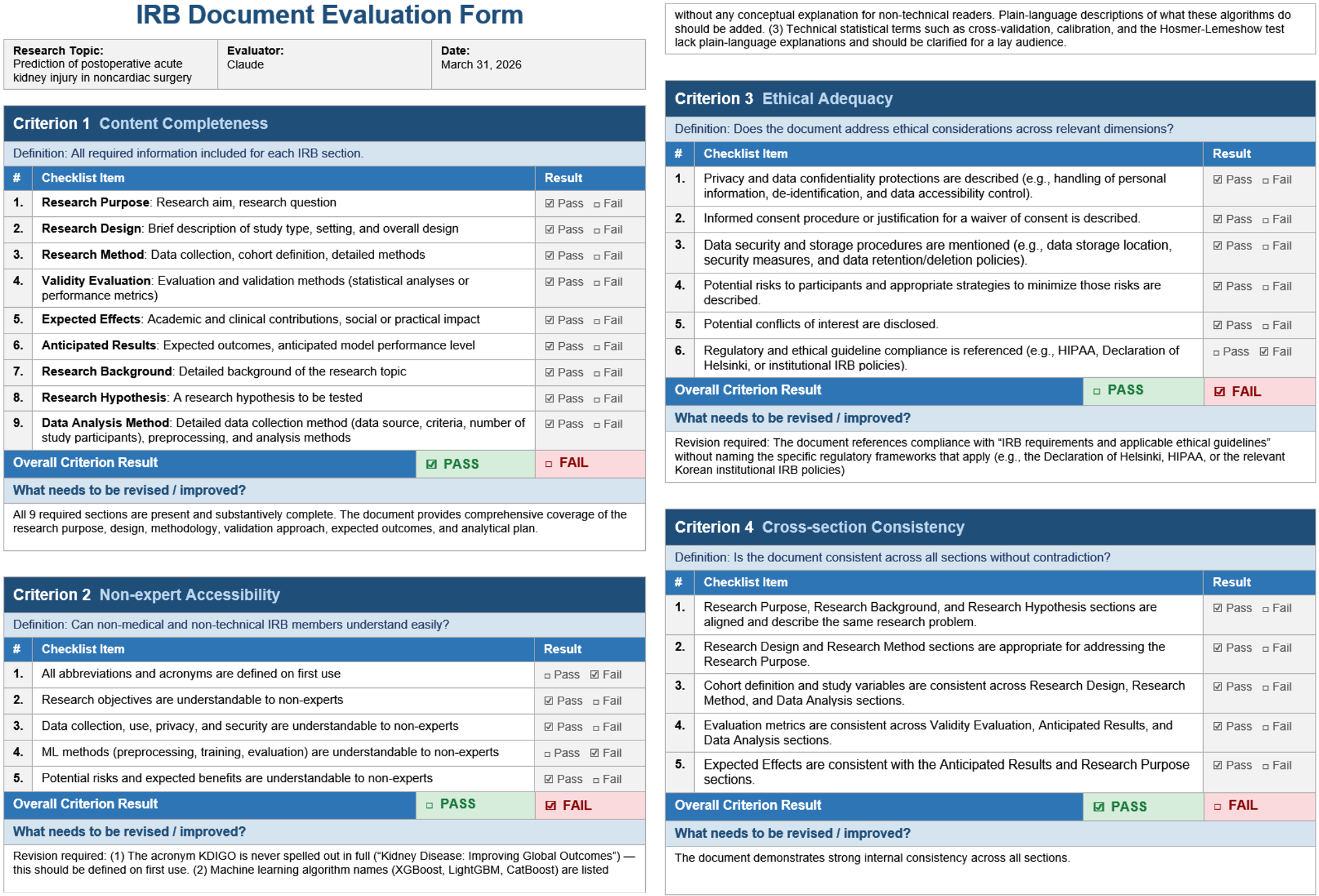}
    \caption{\textbf{IRB document evaluation results on the INSPIRE preoperative AKI prediction task.}}
    \label{inspire_irb_final_evaluation_details}
\end{figure*}

\begin{figure*}[h]
    \centering
    \includegraphics[width=\textwidth]{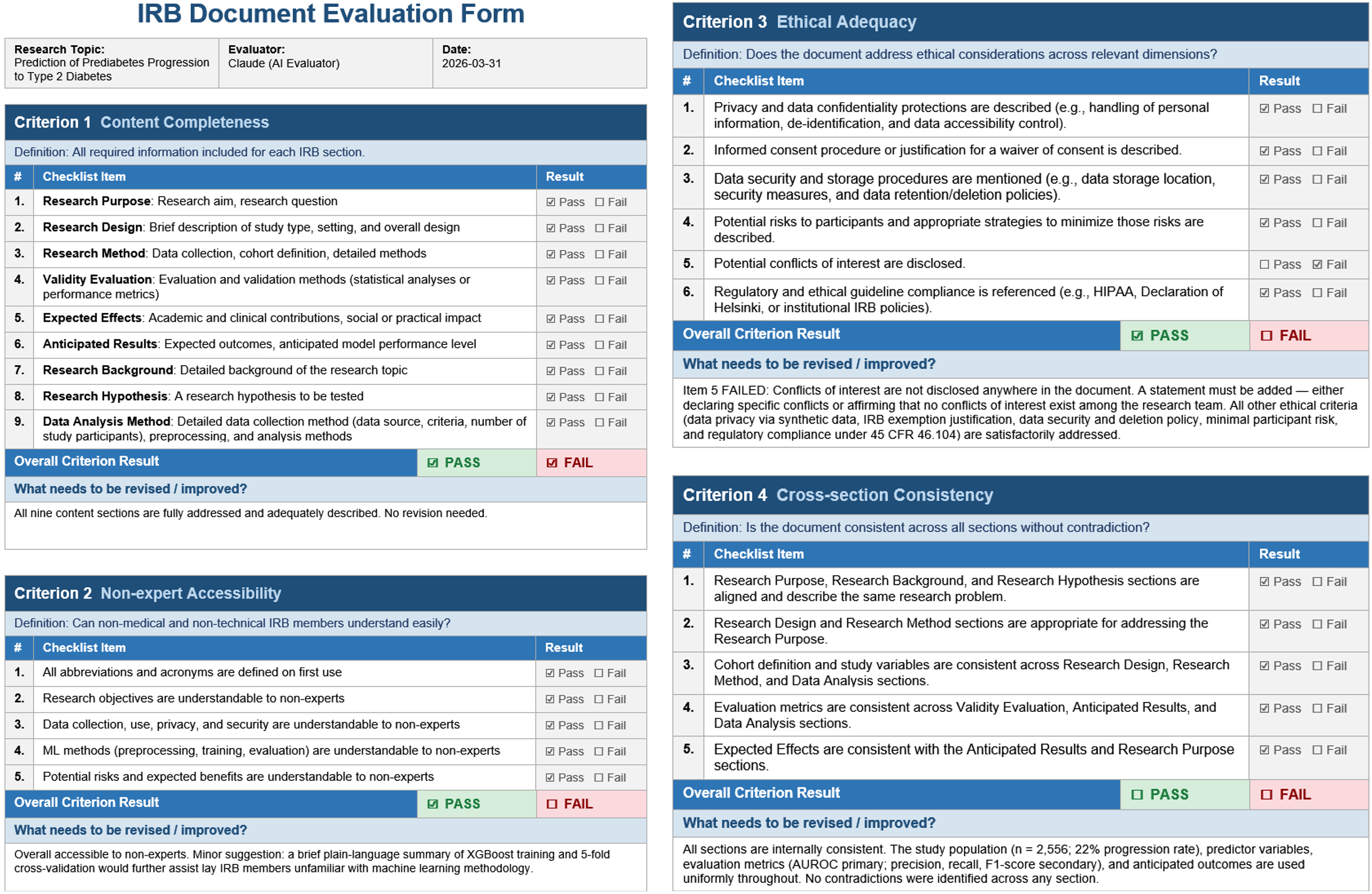}
    \caption{\textbf{IRB document evaluation results on the SyntheticMass ICU readmission prediction task.}}
    \label{syntheticmass_irb_final_evaluation_details}
\end{figure*}

The evaluation format for the final IRB document is shown in Figure S\ref{IRB_document_evaluation_form}. It consists of four criteria, content completeness, non-expert accessibility, ethical adequacy, and cross-section consistency, each defined by a set of checklist items. The results from Claude Sonnet are summarized in Figure S\ref{irb_final_evaluation_details}, with task-specific results presented in Figures S\ref{mimic_irb_final_evaluation_details}, S\ref{inspire_irb_final_evaluation_details}, and S\ref{syntheticmass_irb_final_evaluation_details}.

\clearpage

\subsection*{Supplementary Note 13. Evaluation form on final report.}

 \begin{figure*}[h]
    \centering
    \includegraphics[width=\textwidth]{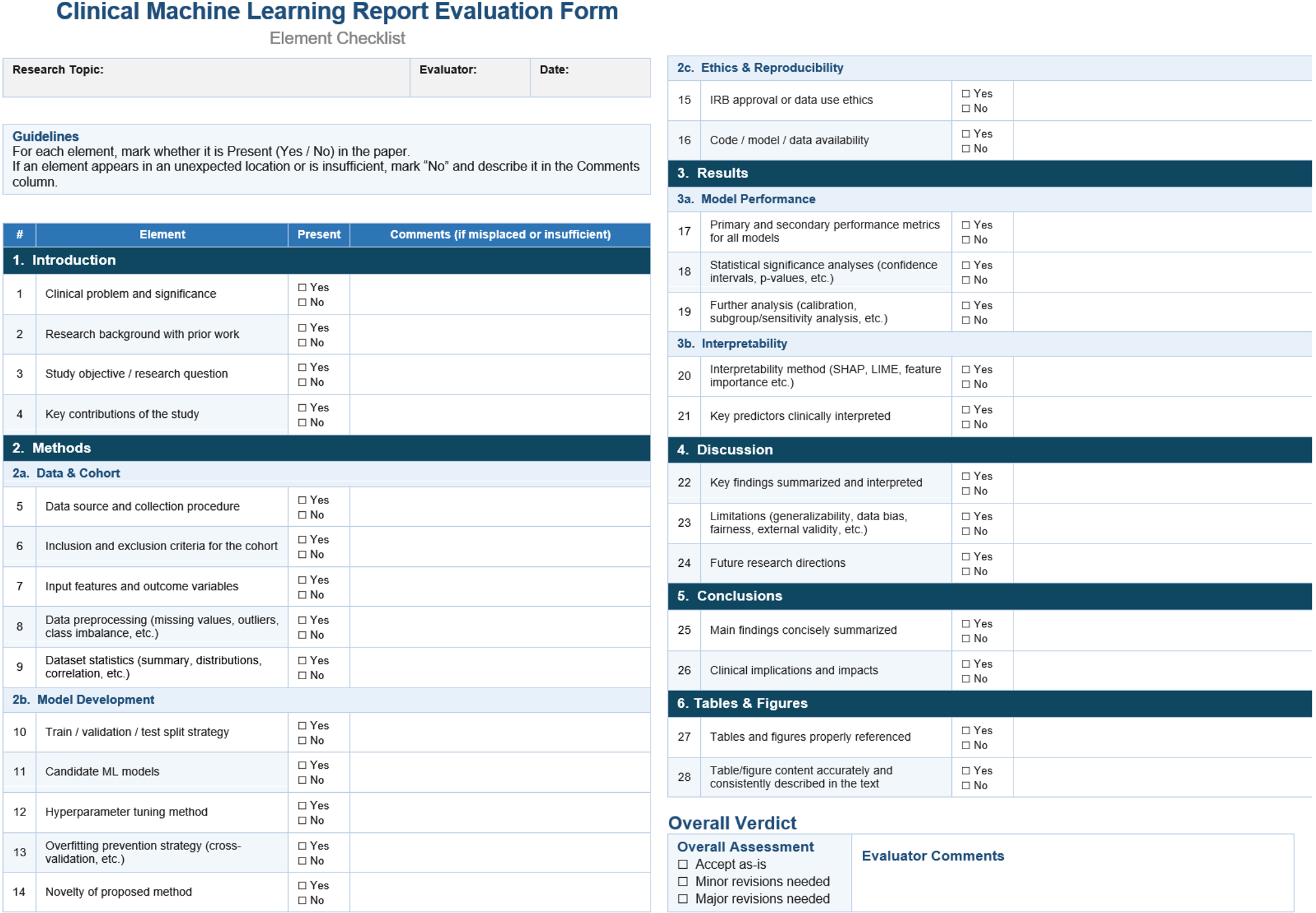}
    \caption{\textbf{Evaluation form used to assess the generated IRB document.}}
    \label{final_report_evaluation_form}
\end{figure*} 

\begin{figure*}[h]
    \centering
    \includegraphics[width=0.8\textwidth]{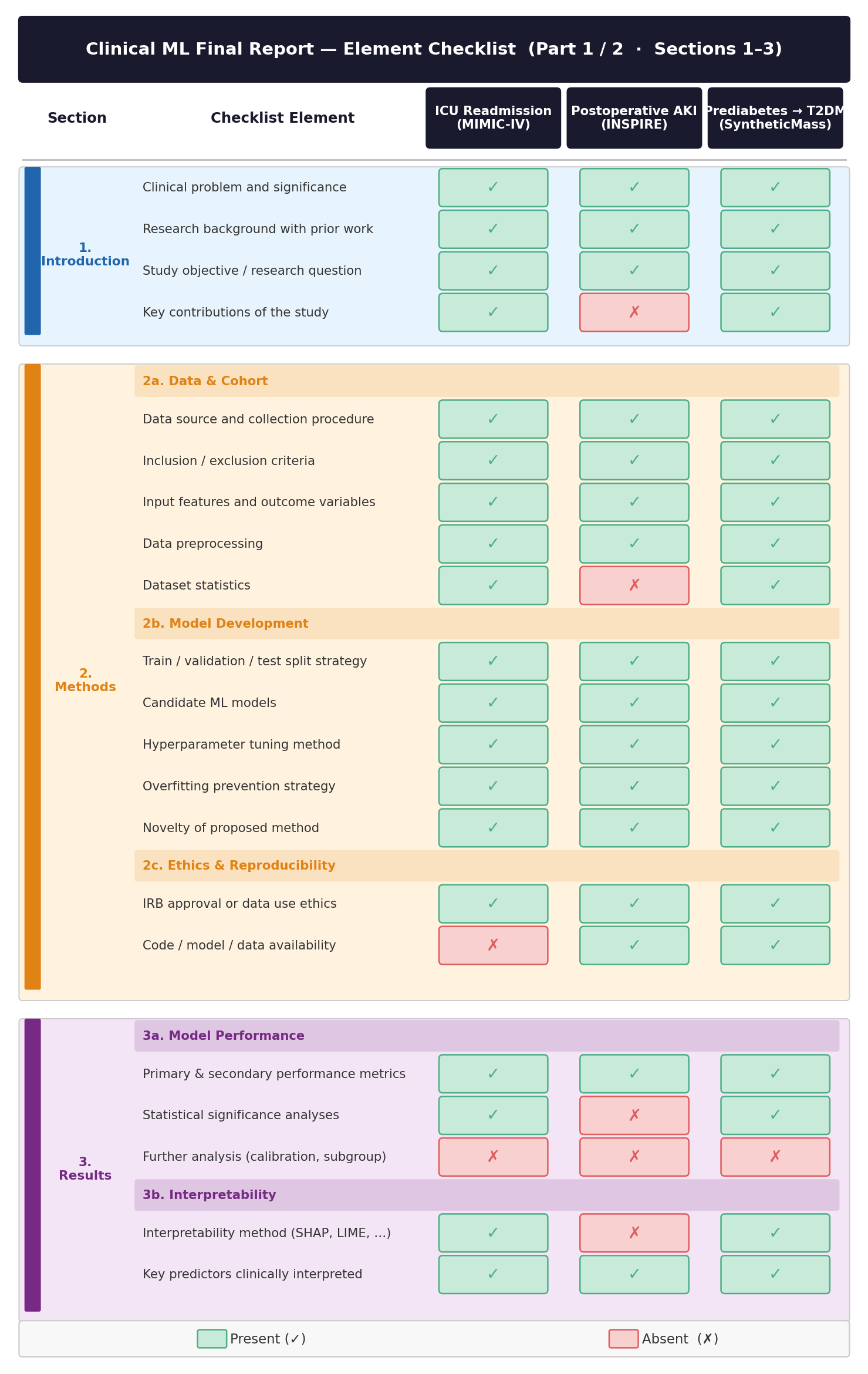}
    \caption{\textbf{Claude Sonnet evaluation results on the final report (section 1-3) across tasks.}}
    \label{final_report_evaluation_details_1}
\end{figure*}

\begin{figure*}[h]
    \centering
    \includegraphics[width=0.8\textwidth]{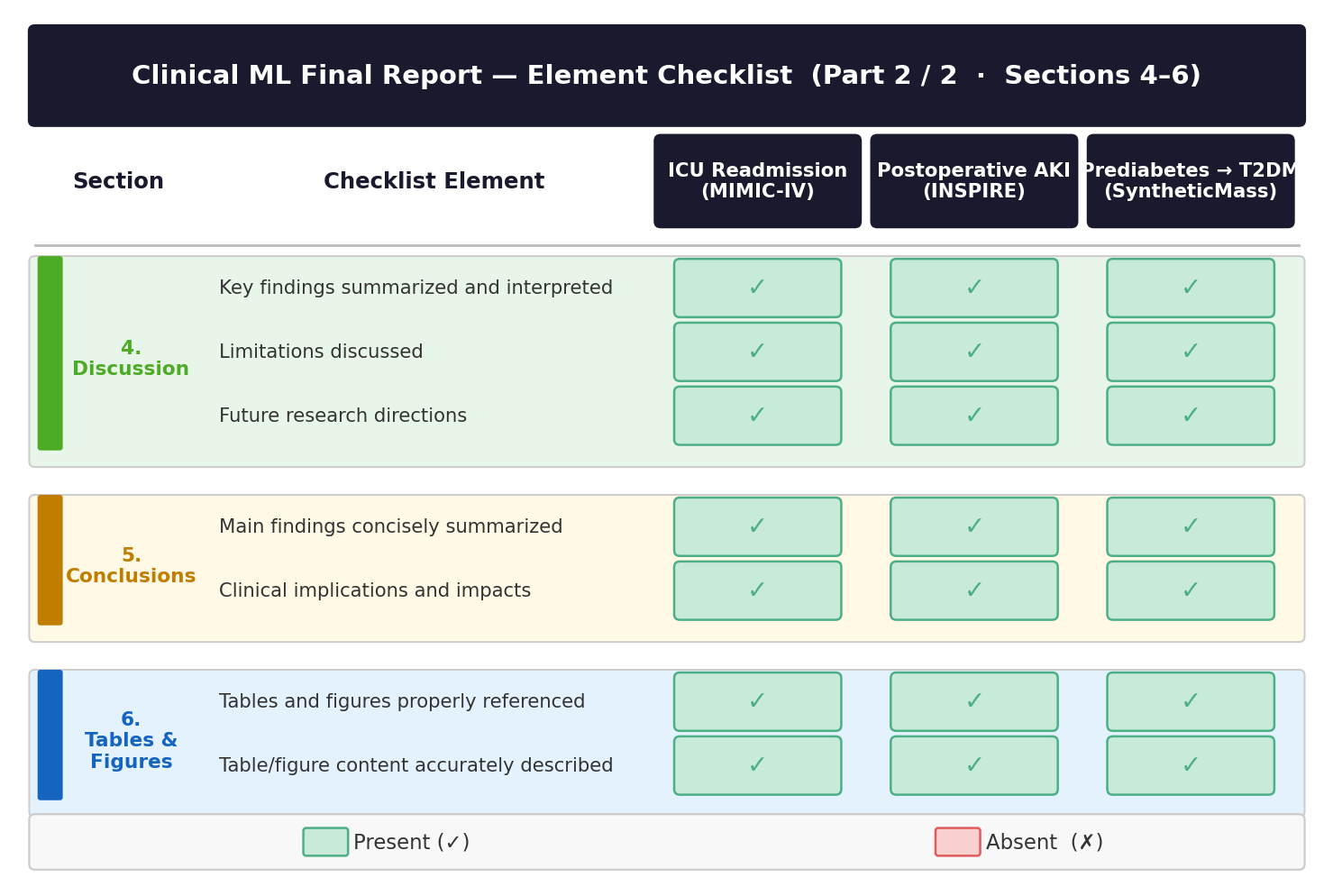}
    \caption{\textbf{Claude Sonnet Evaluation results on the final report (section 4-6) across tasks.}}
    \label{final_report_evaluation_details_2}
\end{figure*}

\begin{figure*}[h]
    \centering
    \includegraphics[width=\textwidth]{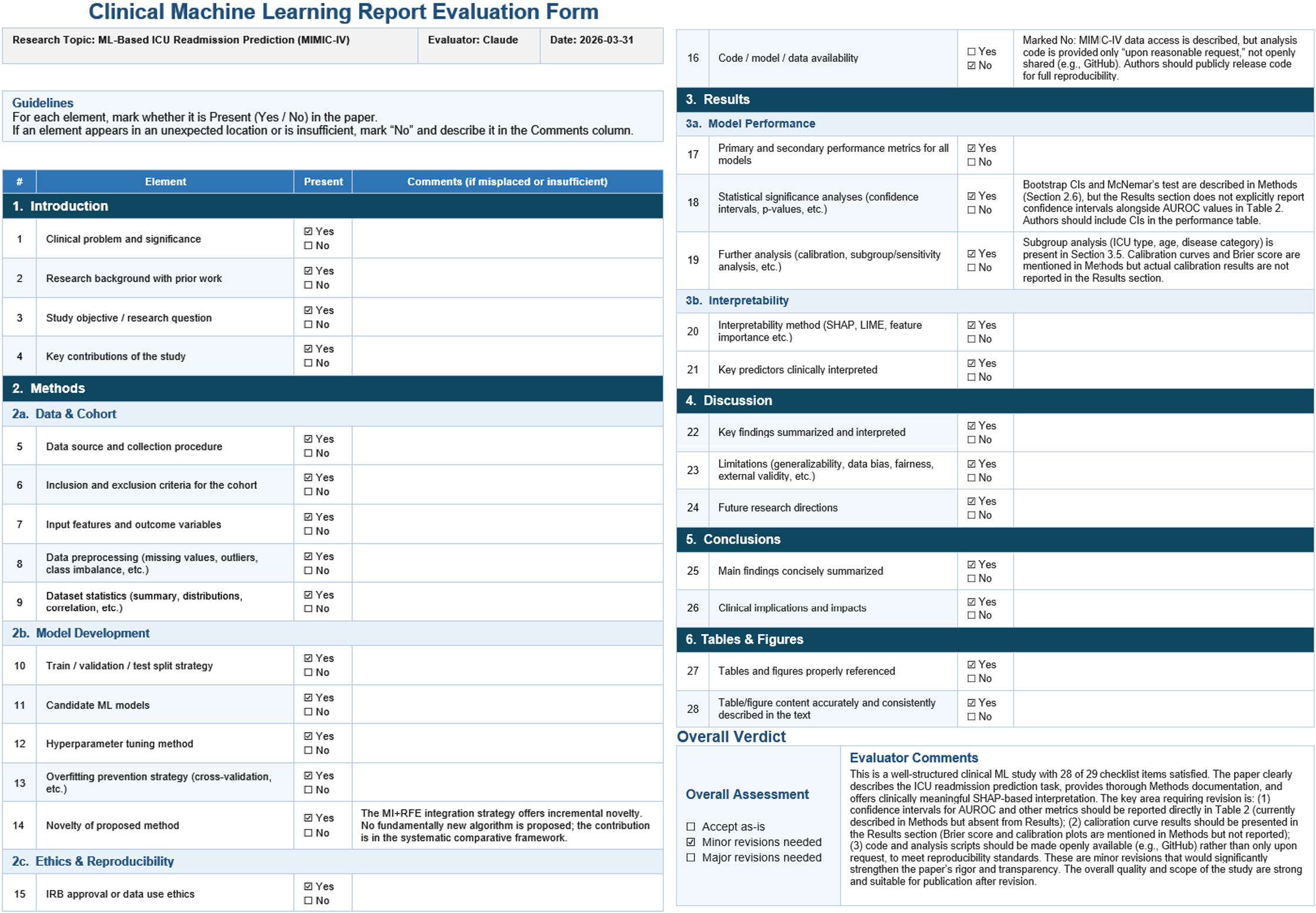}
    \caption{\textbf{Claude Sonnet evaluation results on the MIMIC ICU readmission prediction final report.}}
    \label{mimic_final_report_evaluation_details}
\end{figure*}

\begin{figure*}[h]
    \centering
    \includegraphics[width=\textwidth]{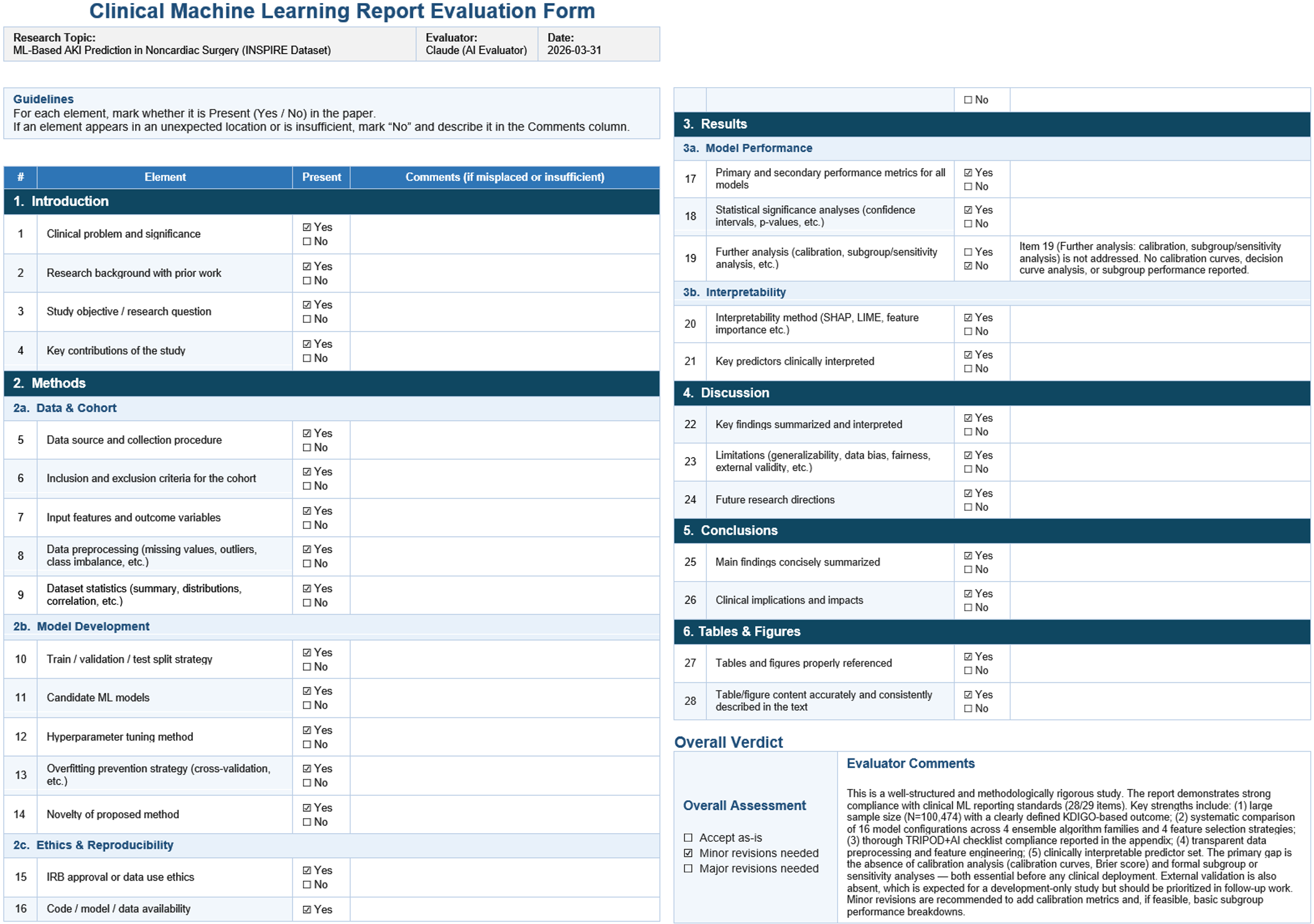}
    \caption{\textbf{Claude Sonnet evaluation results on the INSPIRE preoperative AKI prediction final report.}}
    \label{inspire_final_report_evaluation_details}
\end{figure*}

\begin{figure*}[h]
    \centering
    \includegraphics[width=\textwidth]{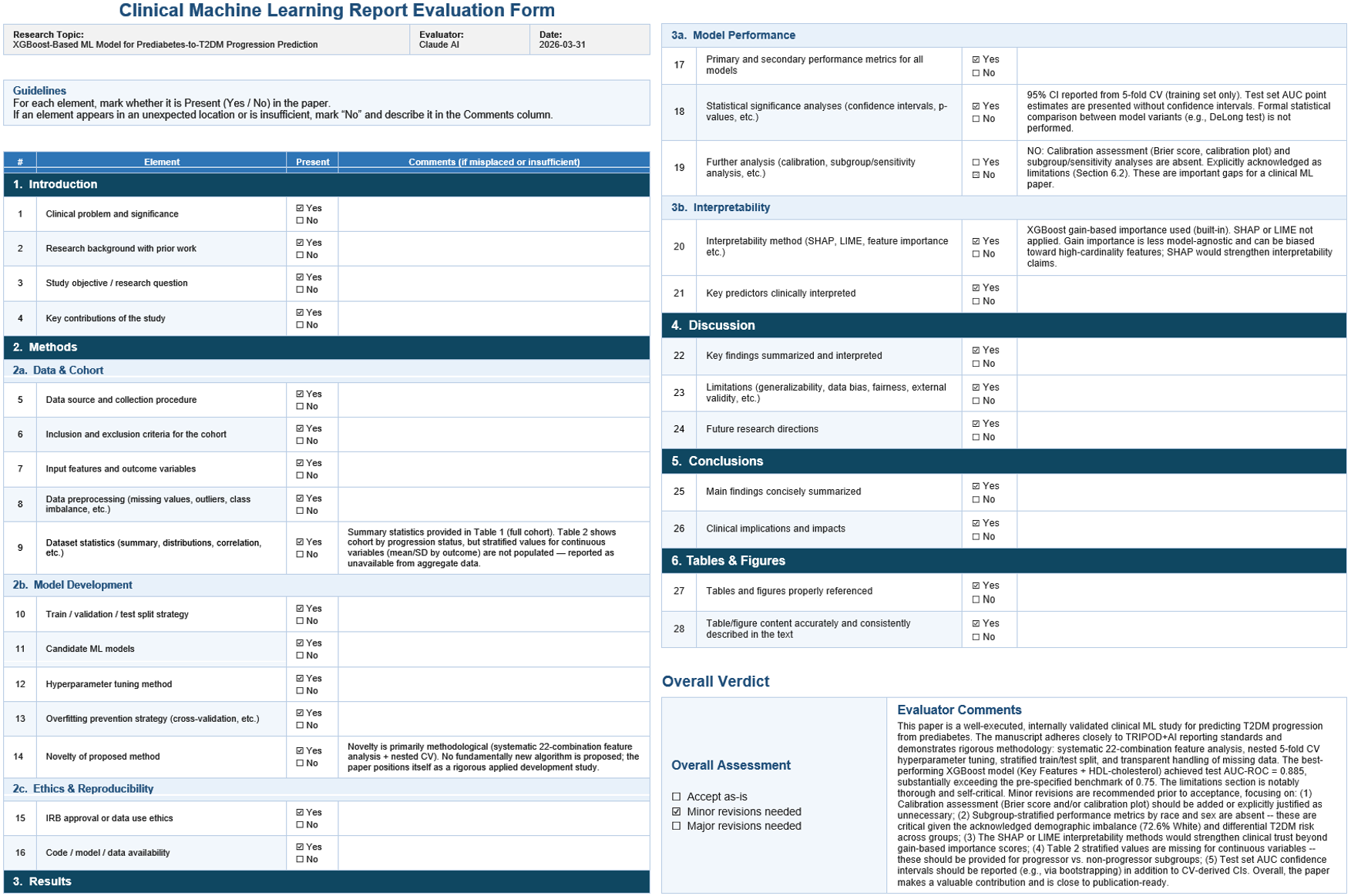}
    \caption{\textbf{Claude Sonnet evaluation results on SyntheticMass ICU readmission prediction final report.}}
    \label{syntheticmass_final_report_evaluation_details}
\end{figure*}

\begin{figure*}[h]
    \centering
    \includegraphics[width=0.8\textwidth]{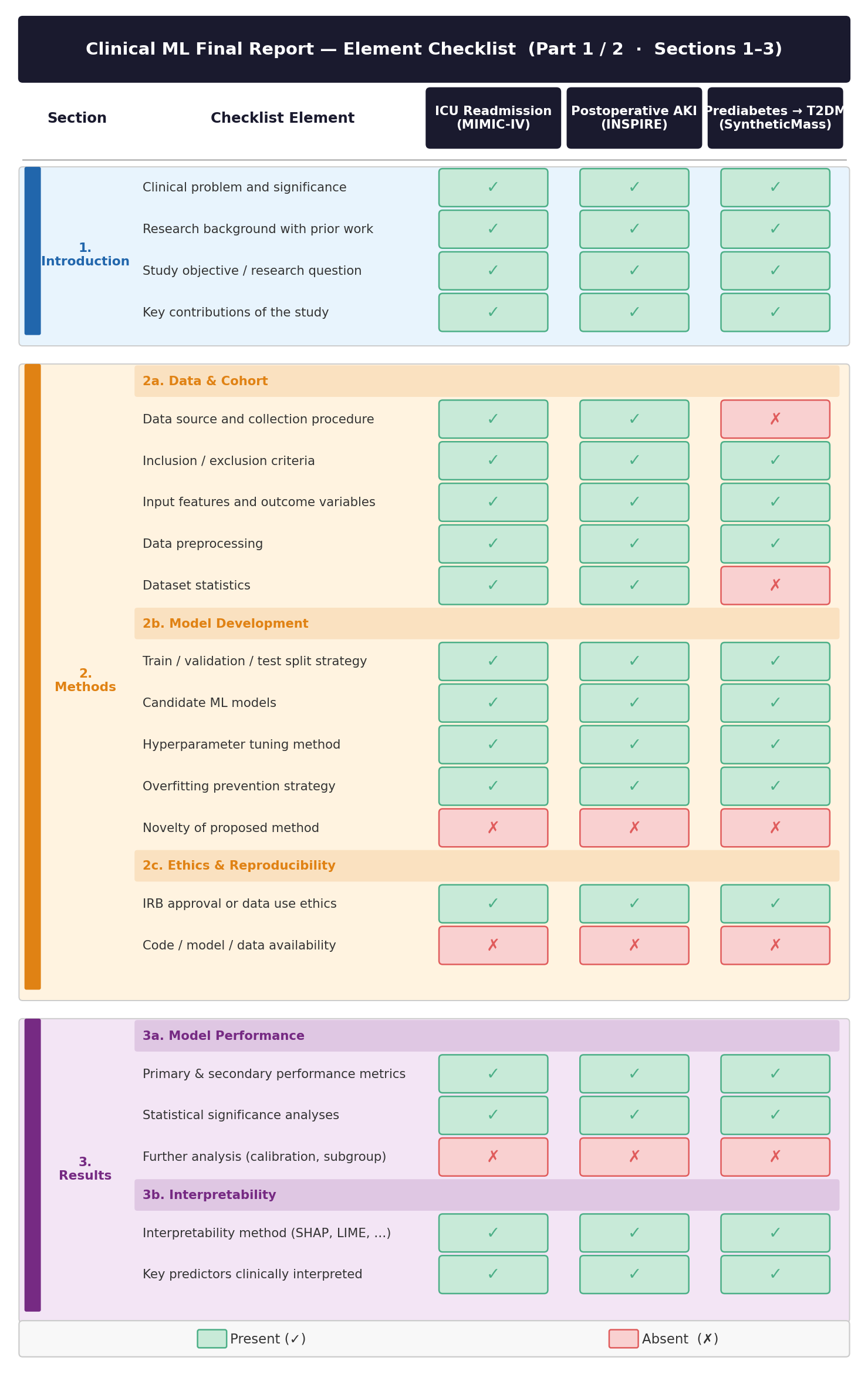}
    \caption{\textbf{Human evaluation results on the final report (section 1-3) across tasks.}}
    \label{final_report_evaluation_details_1_human}
\end{figure*}

\begin{figure*}[h]
    \centering
    \includegraphics[width=0.8\textwidth]{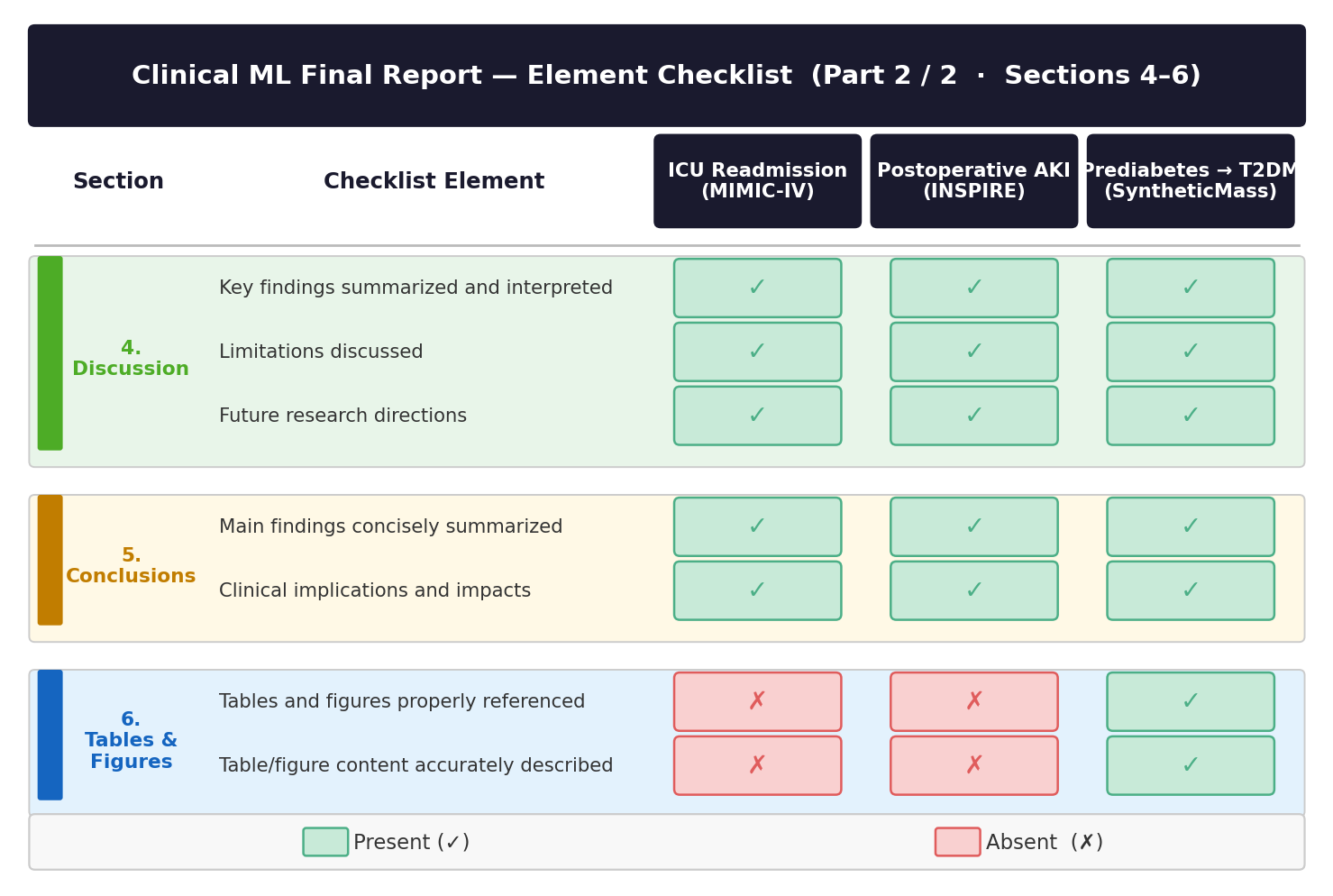}
    \caption{\textbf{Human evaluation results on the final report (section 4-6) across tasks.}}
    \label{final_report_evaluation_details_2_human}
\end{figure*}

\begin{figure*}[h]
    \centering
    \includegraphics[width=\textwidth]{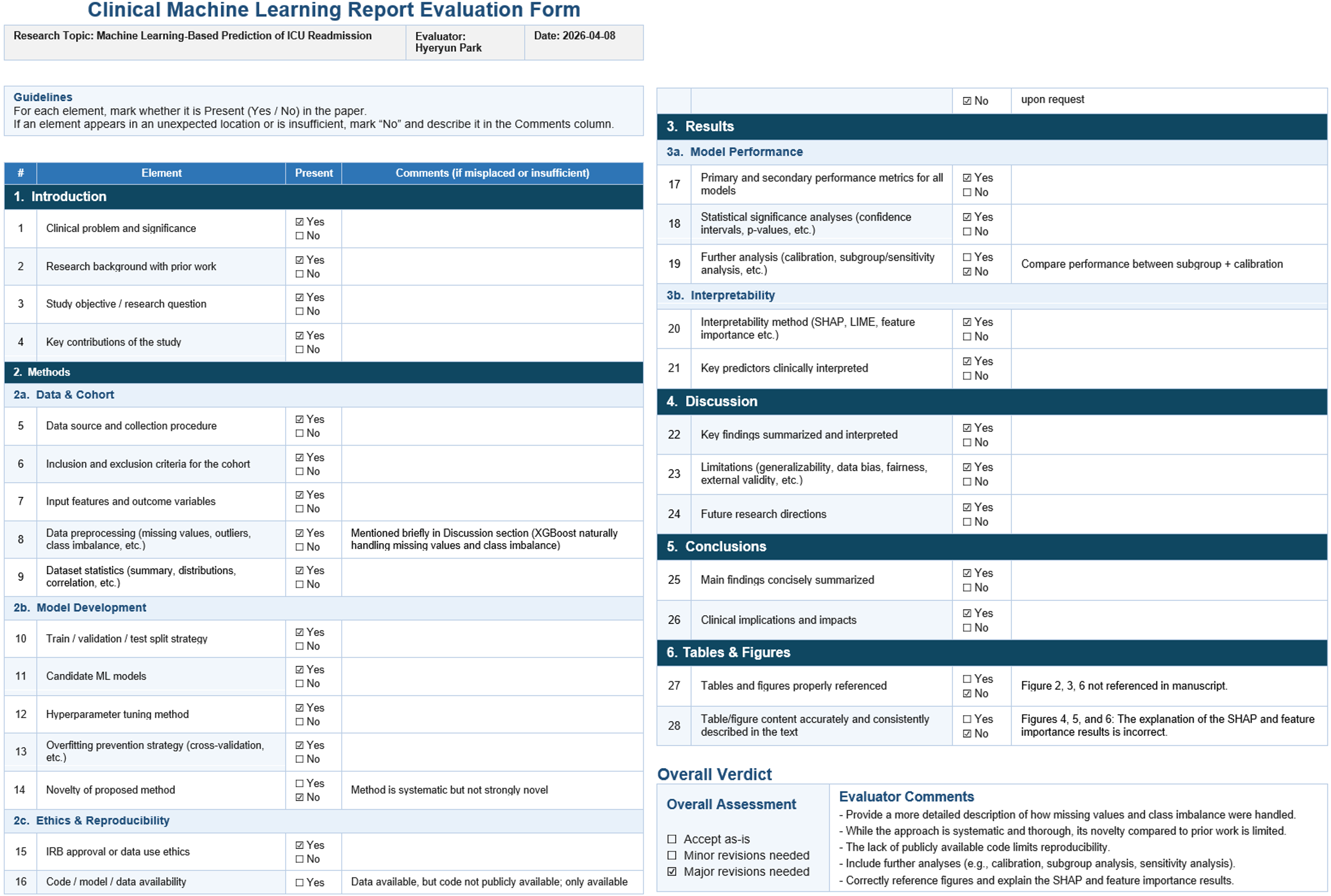}
    \caption{\textbf{Human evaluation results on the MIMIC ICU readmission prediction final report.}}
    \label{mimic_final_report_evaluation_details_human}
\end{figure*}

\begin{figure*}[h]
    \centering
    \includegraphics[width=\textwidth]{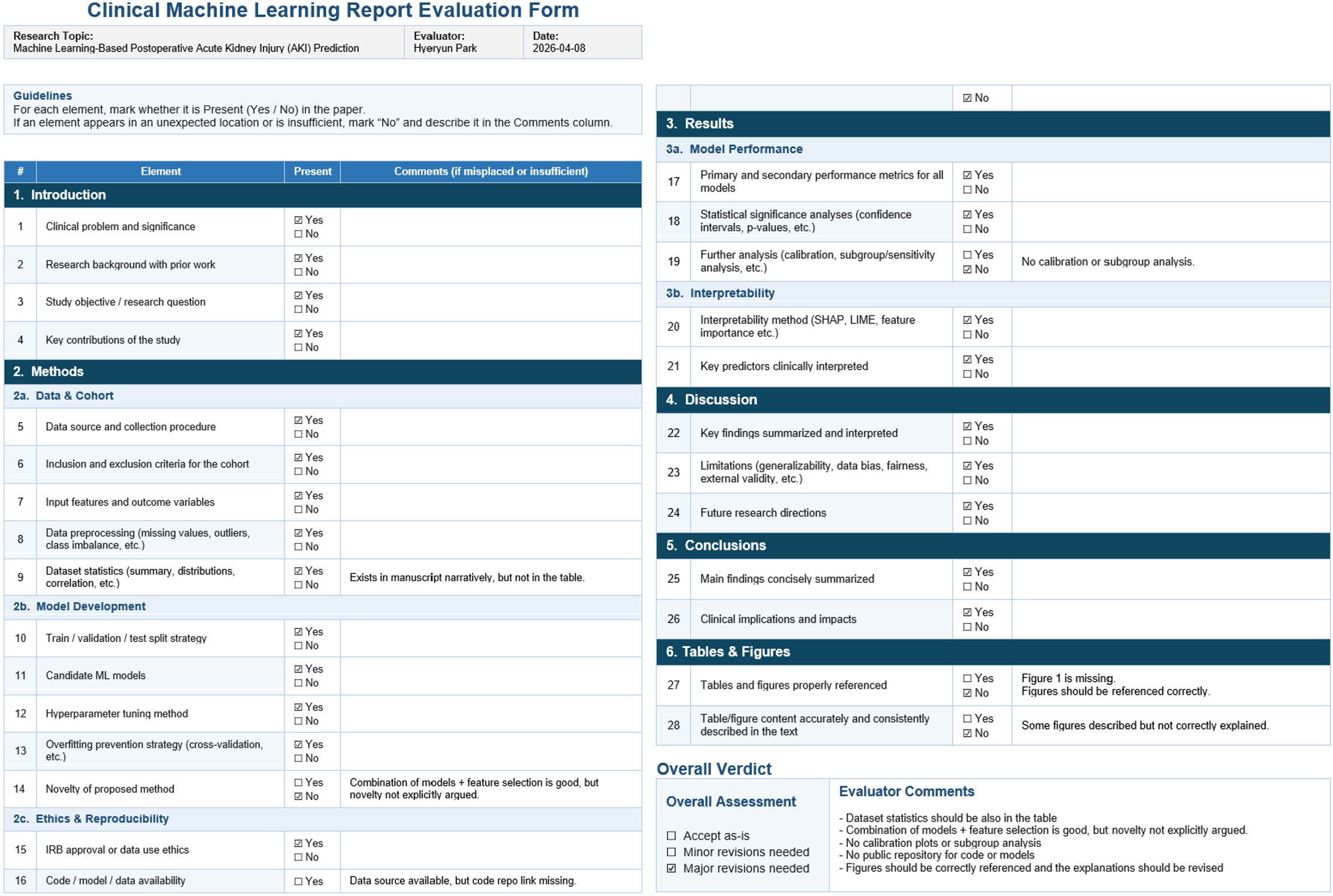}
    \caption{\textbf{Human evaluation results on the INSPIRE preoperative AKI prediction final report.}}
    \label{inspire_final_report_evaluation_details_human}
\end{figure*}

\begin{figure*}[h]
    \centering
    \includegraphics[width=\textwidth]{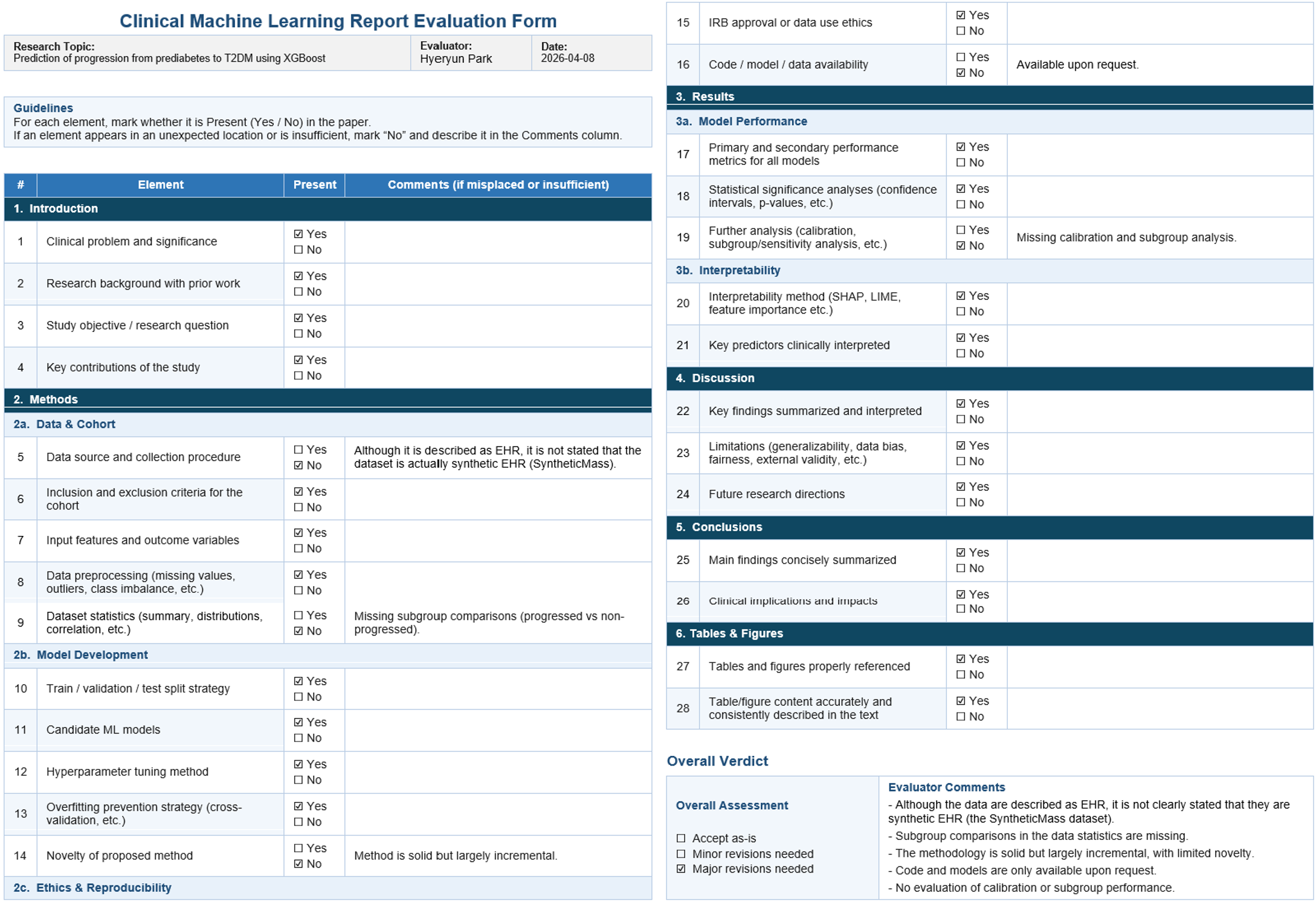}
    \caption{\textbf{Human evaluation results on SyntheticMass ICU readmission prediction final report.}}
    \label{syntheticmass_final_report_evaluation_details_human}
\end{figure*}

As shown in Figure S\ref{final_report_evaluation_form}, a checklist evaluates 28 elements across six sections: Introduction, Methods, Results, Discussion, Conclusions, and Tables \& Figures. The results from Claude Sonnet are summarized in Figure S\ref{final_report_evaluation_details_1} and S\ref{final_report_evaluation_details_2}, with task-specific detailed results presented in Figures S\ref{mimic_final_report_evaluation_details}, S\ref{inspire_final_report_evaluation_details}, and S\ref{syntheticmass_final_report_evaluation_details}.
The results from human evaluation are summarized in Figure S\ref{final_report_evaluation_details_1_human} and S\ref{final_report_evaluation_details_2_human}, with task-specific detailed results presented in Figures S\ref{mimic_final_report_evaluation_details_human}, S\ref{inspire_final_report_evaluation_details_human}, and S\ref{syntheticmass_final_report_evaluation_details_human}.

